\documentclass[10pt,twocolumn,letterpaper]{article}

\usepackage[pagenumbers]{cvpr} % To force page numbers, e.g. for an arXiv version
\usepackage{graphicx}
\usepackage{amsmath}
\usepackage{amssymb}
\usepackage{booktabs}
\usepackage[normalem]{ulem}
\usepackage[ruled,linesnumbered]{algorithm2e}
\usepackage{makecell}
\usepackage{multirow}
\usepackage{color}
\usepackage[table]{xcolor}
\usepackage{etoolbox}
\usepackage{times}
\usepackage{tablefootnote}
\usepackage[numbers]{natbib}

\usepackage{colortbl}
\definecolor{Gray}{gray}{0.9}
\usepackage{epsfig}
\usepackage{float}

\newcommand{\Cmat}[0]{\ensuremath{{\bf C}} }

\newcommand{\Nmat}[0]{\ensuremath{{\bf N}} }

\newcommand{\Xmat}[0]{\ensuremath{{\bf X}} }
\newcommand{\Ymat}[0]{\ensuremath{{\bf Y}} }

\newcommand\blfootnote[1]{%
	\begingroup
	\renewcommand\thefootnote{}\footnote{#1}%
	\addtocounter{footnote}{-1}%
	\endgroup
}

\usepackage[pagebackref,breaklinks,colorlinks]{hyperref}

\usepackage[capitalize]{cleveref}
\crefname{section}{Sec.}{Secs.}
\Crefname{section}{Section}{Sections}
\Crefname{table}{Table}{Tables}
\crefname{table}{Tab.}{Tabs.}

\title{SnapCap: Efficient Snapshot Compressive Video Captioning
}
\author{Jianqiao Sun, Yudi Su, Hao Zhang*, Ziheng Cheng, Zequn Zeng, Zhengjue Wang, Bo Chen*\\
National Key Laboratory of Radar Signal Processing, Xidian University\\
 Xi’an, China, 710071
\\
{\tt\small jianqiaosun@stu.xidian.edu.cn, zhanghao01@xidian.edu.cn, bchen@mail.xidian.edu.cn}
\and
Xin Yuan*\\
Westlake University\\
Hangzhou, China, 310030\\
{\tt\small xyuan@westlake.edu.cn}
}

\begin{document}

\maketitle

\pagestyle{empty}
\thispagestyle{empty}

\blfootnote{* Corresponding authors.
}

\begin{abstract}
\vspace{-5mm}
Video Captioning (VC) is a challenging multi-modal task since it requires describing the scene in language by understanding various and complex videos.
For machines, the traditional VC follows the ``imaging-compression-decoding-and-then-captioning'' pipeline, where compression is pivot for storage and transmission. 
However, in such a pipeline, some potential shortcomings are inevitable, \textit{i.e.}, information redundancy resulting in low efficiency and information loss during the sampling process for captioning. 
To address these problems, in this paper, we propose a novel VC pipeline to generate captions directly from the compressed measurement, which can be captured by a snapshot compressive sensing camera and we dub our model \textbf{SnapCap}.
To be more specific, benefiting from the signal simulation, we have access to obtain abundant measurement-video-annotation data pairs for our model.
Besides, to better extract language-related visual representations from the compressed measurement, we propose to distill the knowledge from videos via a pre-trained CLIP with plentiful language-vision associations to guide the learning of our SnapCap. 
To demonstrate the effectiveness of SnapCap, we conduct experiments on two widely-used VC datasets. 
Both the qualitative and quantitative results verify the superiority of our pipeline over conventional VC pipelines.
In particular, compared to the ``caption-after-reconstruction'' methods, our SnapCap can run at least 3$\times$ faster, and achieve better caption results. 
\end{abstract}

% \vspace{-3mm}
\section{Introduction}
\label{sec: intro}
Video captioning (VC) is an attractive visual-language task, involving understanding dynamic visual contents and generating textual descriptions. 
While describing what we see is a natural task for most people, it is not trivial for machines to do the same~\cite{ramanishka2017top}.
For machines, a straightforward pipeline is ``imaging-compression-reconstruction-and-then-captioning”, as shown in Fig.~\ref{fig: pipeline}(a).
Specifically, a high-definition (HD) video camera captures videos with high resolution in both spatial and temporal domains, which are further compressed for efficient storage and transmission.
% Specifically, high-quality videos with good enough spatial and temporal resolution is captured by camera sensors, and then compressed for  efficient storage and transmission.
Hence, recovering the original video frames is often necessary before generating captions~\cite{zhang2022compressive}.

%Under the traditional imaging framework of “what you see is what you get” with limited resources, most of VC methods usually follow the pipeline of ``imaging-compression-reconstruction-and-captioning”. 
%In other words, starting form the imaging sensor, there is a long way to go before we are ready to do captioning for high-level human understanding.
%Firstly, the high-definition (HD) videos are obtained with hardware and compressed for efficient transmission and storage. 
%Then, the original HD frames are recovered with appropriate methods. 
%Finally, VC models are applied to generate the language descriptions.

\begin{figure}[]
\centering
\includegraphics[width=1.\columnwidth]{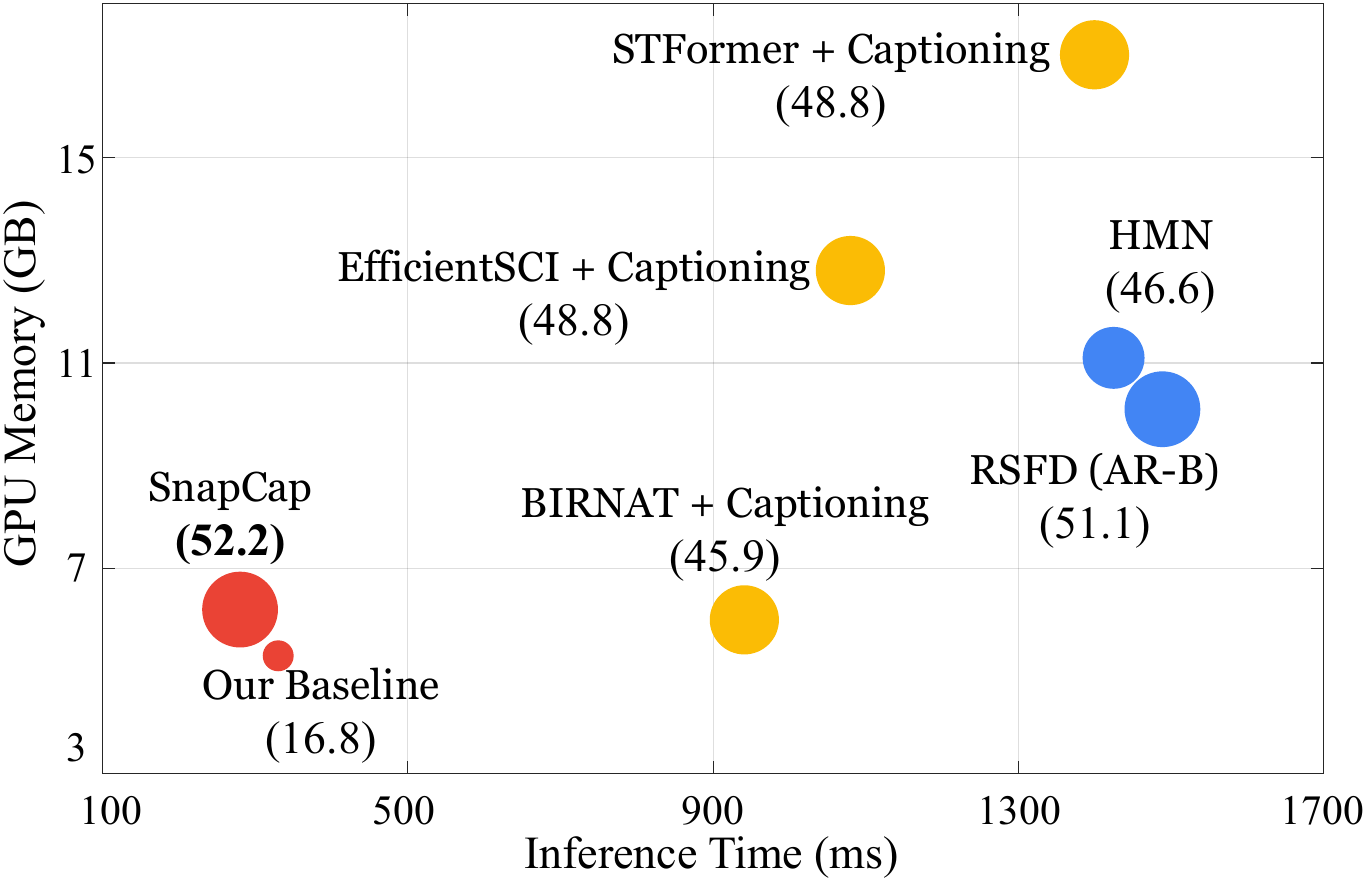}
 \vspace{-4mm}
\caption{Comparisons on GPU memory, inference time, and CIDEr score of typical VC methods, where red, blue, and yellow indicate {\textcolor[RGB]{234,67,53}{our methods}}, {\textcolor[RGB]{66,133,244}{traditional VC methods}}, and {\textcolor[RGB]{251,188,5}{two-stage methods}}, respectively. 
The size of the circle is proportion to the CIDEr score ($\uparrow$) marked in brackets. 
}
\label{fig: comparsion}
\vspace{-4mm}
\end{figure}

Although most VC methods~\cite{yang2021NACF, tang2021clip4caption} assume that they have already obtained the well-decompressed video, they do not consider potential drawbacks of the captioning step in the whole video processing pipeline.
%Though widely adopted [][], this pipeline is tedious and potentially have the following drawbacks:
\textit{i) Information redundancy}: With the increasing spatial and temporal resolutions, the captured raw videos and the reconstructed ones exhibit severe information redundancy, resulting in heavy burden on storage and calculation \cite{cheng2020birnat, stformer, wang2023efficientsci}, as compared in Fig.~\ref{fig: comparsion}.
\textit{ii) Information loss}: To reduce the redundancy in the raw video, (near-)lossless software compression approaches are preferred. However, to handle temporal redundancy in the recovered video, existing VC approaches \cite{SGN, textkg, MGSA} often sample the video frames or video feature maps to reduce computational costs, which in turn may ignore some key information, especially in fast-moving videos.
\textit{iii) Less efficient}: 
As we can see, starting from the captured raw video, there is a long way to go to achieve the output caption, with the help of accumulated efforts of every step. 
However, the redundant information is ``reduced-recovered-and-further-reduced" in the ``compression-reconstruction-and-sampling" loop, which produces a waste of computational resources during the whole pipeline.
% \bb{thus much increasing the computation of the whole pipeline}

%\textit{iv) Poor security}: in security-related scenarios~\cite{secureml}, such as surveillance and military applications, the original frames \rr{are confidential and forbidden to be recovered}.

To realize efficient VC and alleviate computational and storage burden, this paper tries to explore a novel pipeline, describing the scene directly from the data captured by an optical camera, \ie, without {software based} compression nor reconstruction in our way to captioning.
Therefore, there are mainly two questions: \textit{i)} how to efficiently obtain compressive sensed visual data of the live scene; and \textit{ii)} how to build an end-to-end captioning model {directly from the compressive sensed data}. 

%a compressed video measurement.
{To address the aforementioned challenges, we propose to incorporate a typical computational imaging technology~\cite{ci1, ci2}, video snapshot compressive sensing (CS) \cite{sci1, RevSCI}, which physically obtains the compressed measurement during the imaging process.}
% In the recent decades, video snapshot compressive sensing (CS) \cite{sci1, RevSCI}, a typical computational imaging technology \cite{ci1, ci2}, has drawn increasing attention.
Concretely, as shown in Fig.~\ref{fig: SCI}, the optical instrument modulates the live scene via a set of dynamic masks, \eg, produced by digital mirror device (DMD), and then these frames are compressed into a two-dimensional (2D) snapshot measurement by a single exposure of the camera.
%Subsequently, given the measurement and corresponding masks, a wide range of software based algorithms~\cite{desci, gaptv, cheng2020birnat, qiao2020deep, wang2023efficientsci} have been proposed to reconstruct the videos.
Given the measurement, software decoder methods~\cite{desci, gaptv, cheng2020birnat, qiao2020deep, wang2023efficientsci} were proposed to recover the video realistically.
Thus, video snapshot CS enjoys the advantages of low power for imaging sensor, low memory for storage, low bandwidth for transmission, \textit{etc}.~\cite{yuan2014low, zhang2022compressive}.
Therefore, applying the two-stage strategy ``reconstruction-and-then-captioning" (as the yellow pipeline shown in Fig.~\ref{fig: pipeline}) is a potential solution, which still suffers from the low efficiency problem (as the yellow circles shown in Fig.~\ref{fig: comparsion}) similar to traditional VC methods.

% \rr{Although ``sensing-reconstruction-captioniong'', \textit{aka}, the two-stage manner,  is able to reduce the demands for storage compared to the traditional pipeline, as shown in Fig.~\ref{fig: comparsion}, where yellow points are representative methods and..., it still suffers from low efficiency problem where the video frames are necessitated for captioning.}
To overcome this drawback and achieve more efficient VC, we propose an end-to-end approach directly based on the measurement captured by video snapshot CS.
This pipeline is {technically}
%potentially 
feasible, because it is accessible to build supervised data.
Given the masks of the real optical system, we can pretty accurately simulate the acquisition of the measurement (further introduced in Sec.~\ref{sec: SCI}), thus able to build a large-scale training dataset composed by paired measurements, videos, and captions.

% Now, the final challenge is constructing and learning an end-to-end network supervisedly.
The final challenge now is to construct and train an end-to-end network in a supervised manner.
Nevertheless, it is not an easy road, as our previous attempts~\cite{dong2023retrieving, kumawat2022action} discussed in Experiments  (Table.~\ref{Tab:main table} in Sec.~\ref{sec: 4}).
This may be ascribed to the fact that, compared with high-quality videos, the captured measurement is heavily blurred with less visual semantics and moving details, which greatly increases the difficulty to learn effective visual-language representations for caption generation.

To breakthrough these barriers, in this paper, we propose to build a teacher model whose knowledge is distilled to guide the learning of our end-to-end VC network. 
Specifically, as shown in Fig.~\ref{fig: overview}, the teacher model focuses on extracting language-related visual features from the ground-truth video with the help of a pre-trained large vision-language model (VLM), \textit{Contrastive language-image pretraining} (CLIP)~\cite{CLIP}.
Therefore, the teacher model not only conveys spatial and temporal details from the ground-truth video but also provides abundant {prior} knowledge from CLIP. 
With knowledge distillation (KD), the student model is able to reveal a linguistic-related latent representation, which is {injected} into a Transformer decoder to generate the caption.

The main contributions of this paper are as below:
\begin{itemize}

\item{We propose a novel VC pipeline to realize the efficient caption generation, directly from the data captured by video snapshot compressive sensing, { without compression nor reconstruction} in the software processing phase. {This work is also the first attempt at {\em reconstruction-free} VC method based on the video snapshot CS technology.}}

\item{We employ CLIP to construct a teacher model and utilize knowledge distillation to guide the student model to learn language-related visual features, which is further input into a Transformer decoder for caption generation. The whole model is trained in an end-to-end manner.}

\item Comprehensive experimental results on VC benchmarks demonstrate the efficiency and the effectiveness of our SnapCap, which 
{achieves competitive VC scores compared to HD-video-based captioning methods, and run at least 3$\times$ faster compared to two-stage approaches with much better caption results.}
% reduce the memory by more than 50\%, 
%and achieve much better captioning results.
\end{itemize}

\section{Preliminary and Related Works}
\label{sec: related}

\subsection{Video Snapshot Compressive Sensing}
\label{sec: SCI}

\begin{figure}[]
\centering
\includegraphics[width=1.\columnwidth]{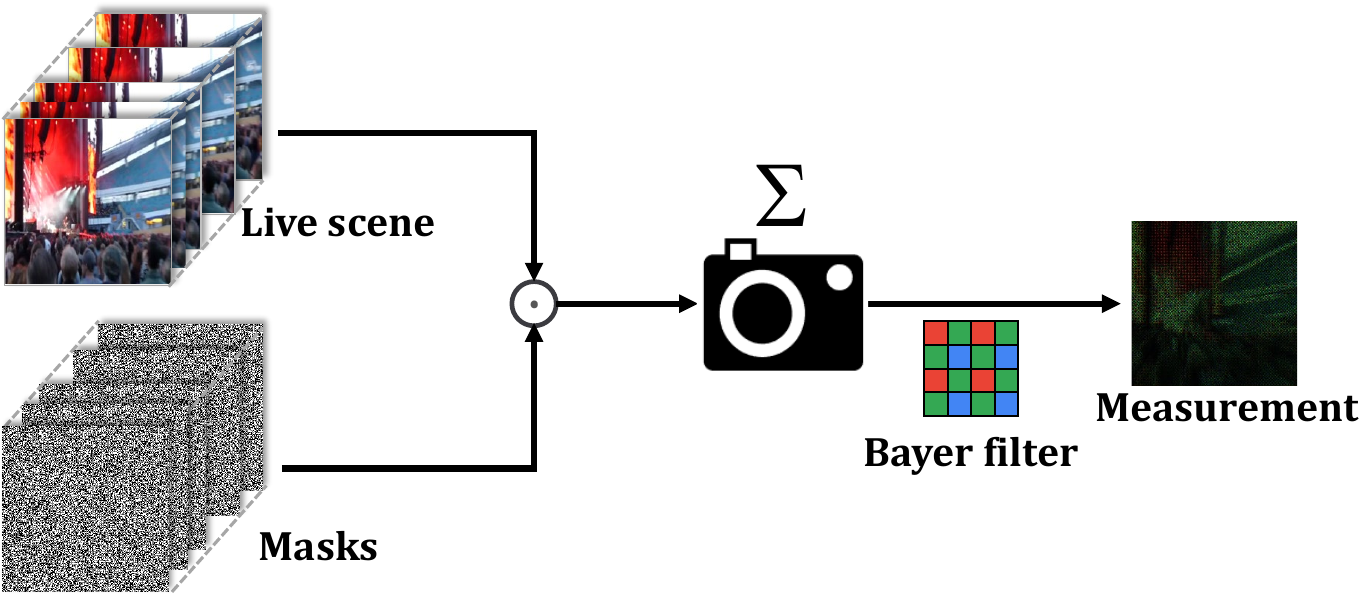}
 \vspace{-6mm}
\caption{Illustration of a video snapshot CS system, CACTI~\cite{cacti}.}
\label{fig: SCI}
 \vspace{-3mm}
\end{figure}

Let's take a typical video CS system, CACTI~\cite{cacti}, as an example.
{As shown in Fig.~\ref{fig: SCI},} we assume the live scene with $B$ high-speed frames $\lbrace{\Xmat_{k}\in\mathbb{R}^{H\times W}}\rbrace_{k=1}^{B}$ is modulated by $B$ coding masks $\lbrace{\Cmat_{k}\in\mathbb{R}^{H \times W}}\rbrace_{k=1}^{B}$.
%of a video are generally modulated by $B$ coding masks $\lbrace{\boldsymbol{C}_{k}}\rbrace_{k=1}^{B}$ ($\boldsymbol{C_{k}}\in\mathbb{R}^{H \times W}$), respectively.
Within one exposure time, the light to the sensor is integrated, thus compressing these coded frames and producing 
a two-dimensional measurement $\Ymat$ via summation, formally as:
 \vspace{-2mm}
\begin{equation}
\label{eq: sci1}
\Ymat=\sum_{k=1}^{B} \Xmat_{k} \odot \Cmat_{k} + \Nmat,
 \vspace{-1mm}
\end{equation}
where $\odot$ and $\Nmat\in\mathbb{R}^{H \times W}$ denote Hadamard (element-wise) product and the noise of the system, respectively. 
{For color video compressive sensing systems, the Bayer filter undergoes spectral sampling before it reaches the sensor. Consequently, considering the linear nature of this process, $\Xmat_{k}$ can be regarded as a mosaic frame.}

% For a single pixel $y_{i,j}$ in the measurement $\boldsymbol{Y}$, it is derived by summing all pixels $x_{i, j, k}$ of the same $(i, j)$ position coded by corresponding mask values $c_{i, j, k}$ in all frames, formulated as:
% \begin{equation}
% y_{i,j}=\sum_{k=1}^{B} c_{i, j, k}x_{i, j, k} + n_{i,j},
% \quad(i=1, ..., H; y=1, ..., W)
% \end{equation}

% For a single pixel $y_{i,j}$ in the measurement $\boldsymbol{Y}$, it is derived by summing all pixels $x_{i, j, k}$ of the same $(i, j)$ position coded by corresponding mask values $c_{i, j, k}$ in all frames.
% Define $\boldsymbol{x}=[\boldsymbol{x}_1^T, ..., \boldsymbol{x}_B^T]$ and $\boldsymbol{\Phi}=[\boldsymbol{D}_1, ..., \boldsymbol{D}_B]$, where $\boldsymbol{x}_k^T=$Vec($\boldsymbol{X}_k$),  $\boldsymbol{D}_k=$Diag(Vec($\boldsymbol{C}_k$)), $k=1, ..., B$, and $\boldsymbol{n}=$Vec($\boldsymbol{N}$); Vec() means the vectorization of the frame or the mask by stacking columns and Diag() is the diagonalization of a vector. Then, the forward coding process can be expressed as
% \begin{equation}
% \label{coding process}
% \boldsymbol{y}=\boldsymbol{\Phi}\boldsymbol{x} + \boldsymbol{n},
% \end{equation}
% where, $\boldsymbol{\Phi} \in \mathbb{R}^{m\times mB}$, the concatenation of $B$ diagonal vectorized masks, is a sparse sensing matrix with $m=HW$ and $\boldsymbol{n}$ denotes to the vectorized noise. 
Therefore, given the coding masks of the real system, one can easily simulate the measurement $\Ymat$ using synthetic data, saving a significant amount of effort
required to capture a large number of real data.
Actually, a training on
simulation and testing on real data framework is widely used in methods developed for recovering the original high-speed frames from the coded measurement \cite{RevSCI, yuan2020plug, wu2021dense, wang2021metasci, li2020end, cheng2020birnat, JalaliY19, qiao2020deep, ma2019deep, wu2023adaptive}.
More introduction to these methods can be found in~\cite{sci1}.

%As shown in \eqref{eq: sci1}, given the coded mask of real system $\{\Cmat_k\}_{k=1}^K$, one can easily simulate the measurement $\Ymat$ with ground-truth video $\Xmat$ (for real system, the noise $\Nmat$ has been considered in mask).
%Therefore, most related research focus on restoring the high-speed frames from the coded measurement with corresponding masks by deep neural networks~\cite{RevSCI, yuan2020plug, wu2021dense, wang2021metasci, li2020end, cheng2020birnat, JalaliY19, qiao2020deep, ma2019deep, wu2023adaptive}.
%More details about these methods can be found in~\cite{sci1}.

%Current research of video SCI focus on the video reconstruction task which aims to restore high-speed frames given the coded measurement and corresponding masks. In the past few years, different algorithms have been proposed for SCI reconstruction~\cite{RevSCI, yuan2020plug, wu2021dense, wang2021metasci, li2020end, cheng2020birnat, JalaliY19, qiao2020deep, ma2019deep, wu2023adaptive}, and more details about these methods can be found in~\cite{sci1}. 

Recently, there is a novel trend towards coupling video snapshot CS with high-level visual understanding tasks, without recovering the original video.
In \cite{hu2021video},
Hu \etal realized video object detection based on the coded measurement directly using a deep CNN network.
For action recognition, Okawara~\textit{et al.}~\cite{kumawat2022action} constructed an end-to-end 3D-CNN model with coded measurement as input.
Both these methods show less complexity and more efficient inferences.
However, their detection/recognition accuracy still falls behind the methods using high-quality video.
Compared with object detection and action recognition, VC is a more challenging task.
Because, besides understanding the visual contents, such as objects or actions, the VC model should also learn visual-language relations for cross-modality generation.
Though challenging, we have achieved comparable  performances with most of existing {HD-video-based VC methods.}
%Moreover, there is still a big performance gap between these methods and those methods using high-quality videos, while our model achieves comparable performance with most of existing VC methods using  high-quality videos.

%Recently, there is a growing trend towards coupling SCI systems and high-level semantic understanding tasks. In~\cite{hu2021video}, the authors introduce the video object detection into the coded image with an optical neural network based encoder, a CNN decoder and an objection detection network. Further, Okawara~\textit{et al.}~\cite{kumawat2022action} propose a reconstruction-free action recognition framework given a single coded measurement and achieves promising performances comparable to traditional 3D-CNN based action recognition network. In this paper, however, we argue that reconstruction has its advantage to capture high-level information from the measurement and proposes a novel semantic-oriented module. 

\subsection{Video Captioning}

In recent years, video captioning has attracted much attention from researchers to understand and describe videos, which can be roughly classified into two groups: attention-based methods and vision-language pretraining-inspired captioning methods. 
In the first group \cite{MGSA, MGRMP, SGN, recnet}, previous works usually employ a 2D or 3D backbone, \textit{e.g.}, ResNet-101~\cite{resnet}, IncepResNetV2~\cite{szegedy2016inceptionv4}, C3D~\cite{C3D}, S3D~\cite{S3D}, to extract spatial and motion features. 
Then, various {characteristic fusion methodologies} are designed and some works also introduce extra information like detection results~\cite{STGKD}, knowledge graph~\cite{textkg} to generate captions.
In the second group \cite{mvgpt, wang2022git, luo2020univl, mplug2}, researchers intend to learn representations between images and texts or videos and texts by first pretraining on large-scale datasets, such as LAION-400M~\cite{schuhmann2021laion}, Howto100M~\cite{miech19howto100m}, and Webvid-2.5M~\cite{bain2021frozen}, and then finetuning the model on downstream tasks and datasets or even perform zero-shot learning~\cite{tewel2022zero}. 
We refer the readers to~\cite{abdar2023review} for more introductions to VC. 
% \rr{[This subsection needs double check!]}

\subsection{Knowledge Distillation}
Knowledge distillation~\cite{gu2021zero,jiao2019tinybert} aims to to transfer knowledge from a complex teacher model to a lightweight student model, which has been widely explored in various applications, such as object detection~\cite{wang2019distilling, zhang2022compressive, kim2020few}, image recognition~\cite{yu2019learning, xu2020feature, ge2021self}, image generation~\cite{wang2018kdgan, li2020gan, li2022learning}, \textit{etc}.
Recently, an increasing number of works focus on using KD to transfer the knowledge from large pre-trained models to domain-specific ones for different tasks~\cite{Chen_2023_ICCV, Wu_2023_ICCV, Chang_2023_ICCV}, achieving {superior}
%the state-of-the-art (SOTA) 
performances than traditional train-from-scratch neural networks.
Except for single-modality knowledge transferring, some researchers also propose to distill the knowledge for cross-modality tasks based on the semantically-abundant data sources~\cite{zhang2023efficient,gupta2016cross, cho2020speech}. 
What we explore in this work is how to transfer the knowledge from the raw data (high-quality video) to the compressed data (coded measurement) via KD technology. 
% Various efforts have been make for model compression, domain transfer 
% It has been widely studied in recent years, with various efforts for exploring different aspects of KD in the context of model compression [] and domain transfer []. 
% Currently, an increasing number of works focus on using KD to transfer the knowledge from a large pre-trained models to domain-specific ones for different tasks, VC included [], which achieves the state-of-the-art performance compared with traditional nerual networks.
%Wang et al. proposed a method called CLIP-TD, which intelligently distills knowledge from both the vision and language branches of CLIP into existing architectures for visual-linguistic tasks. The key idea is that knowledge distillation dynamically adapts per instance by selecting tokens and weighting distillation according to teacher confidence values.
% Basically, these methods usually use KD to transfer the knowledge among the domains of the raw data, while our work explore how to use KD to transfer the knowledge from the raw data (high-quality video) to the compressed data (coded measurement). 

\begin{figure*}[]
\centering
\includegraphics[width=1.0\textwidth]{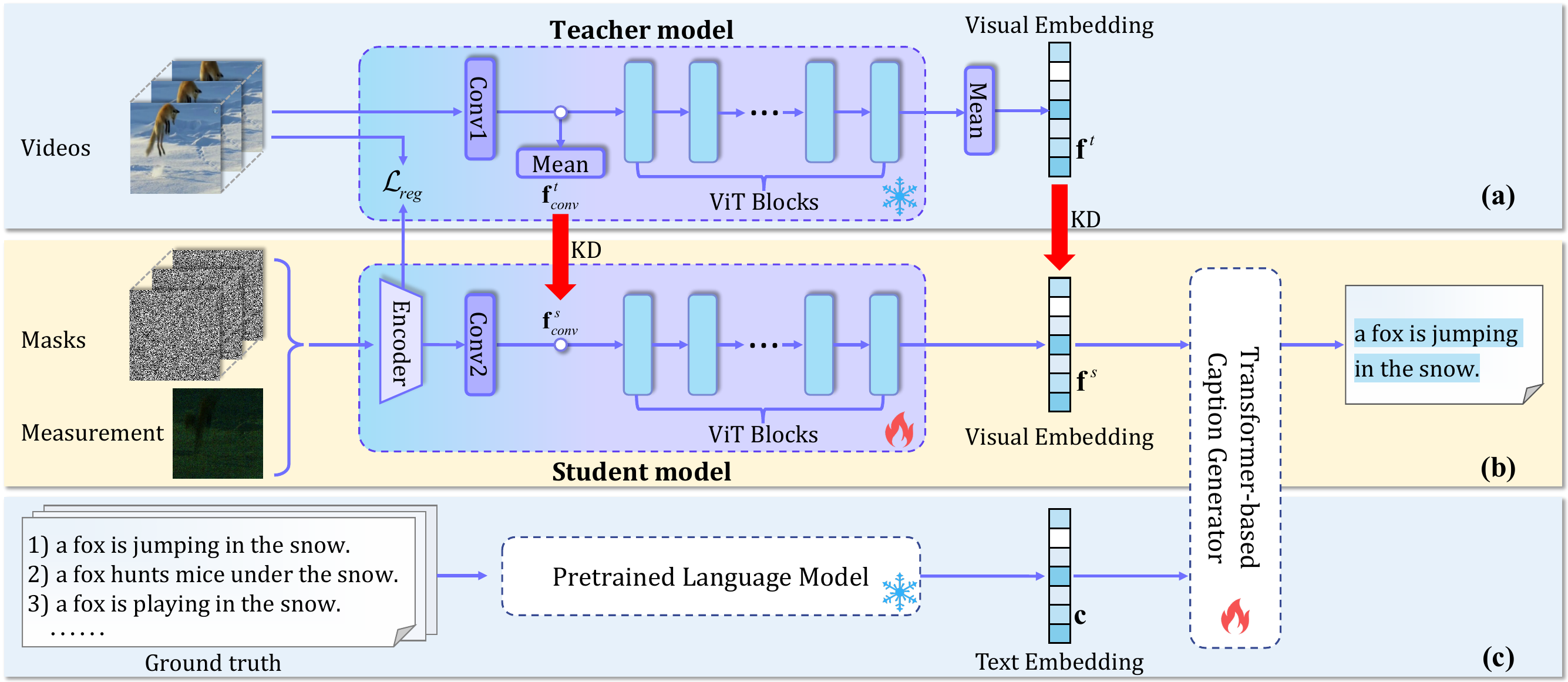}
 \vspace{-6mm}
\caption{Learning and inference workflows of our proposed \textbf{SnapCap}. The cooperation of (a), (b), and (c) is for training, and only (b) is needed for an end-to-end captioning during testing.
%, n overview of our framework, where (a) and (c) are operated during the training process, and (b) is our \textbf{SnapCap}. $\mathcal{L}_{reg}$ is the regularization term, corresponding to Eq.~\eqref{eq: recon loss}.
}
\label{fig: overview}
 \vspace{-3mm}
\end{figure*}

\section{Methodology}
\label{sec: methodology}

To realize efficient captioning {{\em directly from the compressive sensed video snapshot} captured by a computational camera}, we propose a novel video snapshot captioning model, dubbed \textbf{SnapCap}, generating descriptions without compression nor reconstruction. 
{In such a cross-modality generative task, the key is to extract language-related visual features, that are further used for caption generation. Hence, our model consists of a visual extractor and a caption generator, whose structure details as well as the learning and inference details will be introduced below.}
%Hence, in the following subsections, we will introduce our proposed visual extractor, an efficient caption generator, as well as the learning and inference details. 

\subsection{Visual Encoder via Knowledge Distillation}
\label{sec: encoder}
Given a compressed measurement $\Ymat$ and its corresponding masks $\{\Cmat_k\}_{k=1}^B$ shown in Fig.~\ref{fig: overview} (b), a straightforward method to obtain textual predictions is to train a captioning model like most VC methods~\cite{recnet, MGSA, STGKD} and then perform inference. However, owing to the fact that the compressed data $\Ymat$ is always heavily blurry and noisy with much fewer details than HD video frames, such a direct manner fails to yield satisfactory results~\cite{dong2023retrieving, kumawat2022action}, and it is a very challenging task to capture effective visual features {(as our previous attempts discussed in Experiments)}. 
Thanks to the accessible simulation (as introduced in Sec.~\ref{sec: related}), we can obtain abundant video-measurement data pairs and distill the knowledge from the video to the measurement. 
Hence, we hope to build a teacher model to capture effective visual information from the ground-truth video, which can be employed to guide the feature extraction from the measurement, \ie, the student model $S(\cdot)$.

Specifically, considering the vision-language association knowledge incorporated in the pre-trained model CLIP~\cite{CLIP}, which is trained on a large-scale image-text pairs~\cite{schuhmann2021laion}, we apply the image encoder of the CLIP to capture the information contained in the video, which is denoted as the teacher model $T(\cdot)$. 
Nevertheless, given that there is a large discrepancy between the inputs of the teacher and the student models, it is infeasible to directly copy the structure of the teacher model to the student one.
To solve this problem, we propose to map the video and the measurement to a shared latent space.

% While during the production of low-information-density $\Ymat$, the original high-quality video frames $\{\Xmat_k\}_{k=1}^B$ are involved, which inspires us to alleviate the issue with the help of knowledge distillation strategy. To be more specific, focusing on extracting language-relevant visual features, we propose to employ the CLIP model~\cite{CLIP} as our teacher model $T()$\footnote{In our work, only the image encoder part of the CLIP is involved.}, which is trained on a large-scale image-text pairs~\cite{schuhmann2021laion}. In this manner, the information and the knowledge contained in the video can be captured effectually. 
To be more specific, we transform the videos to the first convolutional layer $\mathrm{Conv1}(\cdot)$ of the CLIP image encoder to get the feature maps in an efficient manner as:
\begin{equation}
\boldsymbol{\mathrm{f}}_{conv}^t=\mathrm{Mean}(\mathrm{Conv1}(\Xmat_1, ..., \Xmat_B)) \in \mathbb{R}^{c\times h\times w},
\end{equation}
where $\mathrm{Mean}(\cdot)$ denotes the average pooling operation. Given the measurement $\Ymat$ with much blur and less details, a single-layer convolutional operation is hard to extract meaningful semantics to match the $\boldsymbol{\mathrm{f}}_{conv}^t$. 
{Thus, we introduce an encoder $f(\cdot, \cdot)$ consisted of multiple residual blocks to extract the latent representation from the measurement},
\begin{equation}
\boldsymbol{\mathrm{f}}_{latent}=f(\Ymat, \Cmat).
\end{equation}
Then considering that different CLIP structures contain various parameter settings, we include a two-layer flexible convolutional operation $\mathrm{Conv2}(\cdot)$ after the encoder $f(\cdot, \cdot)$ for feature map alignment,
\begin{equation}
\boldsymbol{\mathrm{f}}_{conv}^s=\mathrm{Conv2}(\boldsymbol{\mathrm{f}}_{latent}), \boldsymbol{\mathrm{f}}_{conv}^s\in \mathbb{R}^{c\times h\times w},
\end{equation}
In this manner, the feature maps of the video and the measurement can be extracted into a shared latent space. 
Besides, the follow-up structure of the teacher model, \eg, the ``ViT blocks'' in Fig.~\ref{fig: overview}, can be copied to student model as an initialization. 
With $\boldsymbol{\mathrm{f}}_{conv}^t$ and $\boldsymbol{\mathrm{f}}_{conv}^s$ in the same dimension and holding similar semantic representations, we further extract language-related vision embeddings for the video and the measurement, respectively, which can be formulated as:
\begin{align}
\boldsymbol{\mathrm{f}}^t&=\mathrm{Mean}(T(\Xmat_1, ..., \Xmat_B))\in \mathbb{R}^{d},\\
\boldsymbol{\mathrm{f}}^s&=S(\Ymat, \Cmat)\in \mathbb{R}^{d}.
\end{align}
With such an efficient design, the abundant semantic information embodied in the video can be distilled to the measurement. 
Hence the distillation loss between the teacher model and student model can be written as:
\begin{align}
\label{eq: conv loss}
\mathcal{L}_{conv}&=\mathcal{L}_{MSE}(\boldsymbol{\mathrm{f}}_{conv}^s, \boldsymbol{\mathrm{f}}_{conv}^t),\\
\label{eq: embedding loss}
\mathcal{L}_{emb}&=\mathcal{L}_{MSE}(\boldsymbol{\mathrm{f}}^s, \boldsymbol{\mathrm{f}}^t),\\
\mathcal{L}_{dis}&=\mathcal{L}_{conv} + \alpha \mathcal{L}_{emb},
\end{align}
where $\mathcal{L}_{MSE}$ is the mean-square-error distance between two terms and $\alpha$ is a coefficient. 

In addition to distilling the knowledge from the videos through the direct feature map alignment, treating the video as a regularization term can also help $f(\cdot, \cdot)$ and $\mathrm{Conv}2(\cdot)$ to extract coherent semantics from the blurry measurement~\cite{desci}. 
To this end, we design an efficient decoder $g(\cdot)$, which maintains the systematic network architecture as $f(\cdot, \cdot)$ to recover videos from the latent representation $\boldsymbol{\mathrm{f}}_{latent}$ so that both the spatial and temporal details from the video can be conveyed to the measurement, formulated as:
\begin{equation}
\label{eq: recon loss}
\hat{\Xmat}=g(f(\Ymat, \Cmat)),\\
\mathcal{L}_{reg}=\Sigma \mathcal{L}_1(\hat{\Xmat}, \Xmat).
\end{equation}

Both the distillation loss and the regularization term can help the student model to fully absorb the knowledge from teacher model and obtain meaningful vision embeddings for captioning (verified by our experiments in Sec.~\ref{sec: regularization ablation}).

\subsection{Caption Generator}
After extracting the language-related visual representation $\boldsymbol{\mathrm{f}}^s$ from the student model $S(\cdot)$, we design a lightweight projector $\mathrm{Proj}(\cdot)$ to map the vision embedding to the text space,
\begin{equation}
\label{eq: proj}
\boldsymbol{\mathrm{t}}=\mathrm{Proj}(\boldsymbol{\mathrm{f}^s}), \boldsymbol{\mathrm{t}}\in \mathbb{R}^{D}, 
\end{equation}
where $D$ is the dimension of the text embedding space. At the position $i$ of the sentence, the word can be generated as:
\begin{align}
\boldsymbol{\mathrm{c}}_{< i}&=\mathrm{PLM}(y_{< i}),\\
\boldsymbol{\mathrm{z}_i}&=\mathrm{Concat}(\boldsymbol{\mathrm{t}}, \boldsymbol{\mathrm{c}}_{< i}),\\
p(Y_i)&=\mathrm{Dec}(\boldsymbol{\mathrm{z}_i}),
\end{align}
where $y_{< i}$ is the generated words before the position $i$, $\mathrm{PLM}(\cdot)$ means a Pre-trained Language Model (PLM) such as BERT~\cite{devlin2018bert} to convey the words into the embedding space, $\mathrm{Concat}(\cdot, \cdot)$ is concatenation, and $\mathrm{Dec}(\cdot)$ is a Transformer-based language decoder to generate $y_i$.  

\subsection{Learning and Inference}
\label{sec: loss}

During training, given the original frames, we distill the knowledge from video domain to the blurry coded measurement domain via two objectives, which are treating the video as a regularization term as $\mathcal{L}_r$ and transfer the knowledge incorporated in teacher model through the distillation process $\mathcal{L}_{conv}$ and $\mathcal{L}_{last}$. 
{Following \cite{SGN, ye2022hierarchical, recnet}, }
Given the ground truth annotations $Y^{*}_{1:L}$, as in most previous VC works~\cite{SGN, ye2022hierarchical, recnet}, we adopt the cross-entropy loss to supervise the learning process:
\begin{equation}
\label{eq: cap loss}
\mathcal{L}_{cap}=-\sum_{i=1}^{L} \log p(y_i^{*}|\boldsymbol{\mathrm{f}}^s, y_{<i}^{*}), 
\end{equation}
where $L$ is the length of prediction. 

Take a step further, considering that the optimization objective of including the videos as a regularization term is not exactly the same as performing feature map alignment, directly optimizing the parameters via the combining loss may bring about the convergence issue. 
To mitigate it, and inspired by masked auto-encoder (MAE)~\cite{he2022masked}, we propose to optimize the encoder $f(\cdot, \cdot)$ and the decoder $g(\cdot)$ through  $\mathcal{L}_r$ firstly. Then, without the involvement of $g(\cdot)$, we update the parameters of encoder $f(\cdot, \cdot)$, student model $S(\cdot)$, and projector $\mathrm{Proj}(\cdot)$ through the loss function:
\begin{equation}
\label{eq: total loss}
\mathcal{L}_{total} = \mathcal{L}_{dis} + \beta \mathcal{L}_{cap}, 
\end{equation}
where $\beta$ is another coefficient. 
As suggested by previous works, we employ  a language model to perform as the decoder $\mathrm{Dec}(\cdot)$, where the parameters are frozen to reduce the training complexity.

As shown in Fig.~\ref{fig: overview} (b), during the inference process where only the coded measurement $\Ymat$ and masks $\Cmat$ are given, we input them to the encoder and the student model to perform the forward mapping and derive the language-related vision embedding as:
\begin{equation}
\boldsymbol{\mathrm{f}}^s=S(\Ymat, \Cmat). 
\end{equation}
Then the predicted caption is generated in an auto-regressive word-by-word manner.

The detailed network structure of the our model, the training and inference algorithms can be found in the Appendix. 

\begin{table*}[t]
\centering
\begin{tabular}{lccccccccc}
\toprule
\multirow{2}{*}{Methods} & \multirow{2}{*}{Input modalities} & \multicolumn{4}{c}{MSRVTT~\cite{msrvtt}} & \multicolumn{4}{c}{MSVD~\cite{msvd}} \\
\cmidrule(r){3-6} \cmidrule(r){7-10}
& & B$\uparrow$ & M$\uparrow$ & R$\uparrow$ & C$\uparrow$ & B$\uparrow$ & M$\uparrow$ & R$\uparrow$ & C$\uparrow$ \\
\midrule
\multicolumn{4}{l}{\textbf{\textit{Video frames based methods}}} & \\
RecNet~\cite{recnet} & Vision & 39.1 & 26.6 & 59.3 & 42.7 & 52.3 & 34.1 & 69.8 & 80.3 \\
MGSA~\cite{MGSA} & Vision & 41.7 &27.5 & - & 48.1 &\textbf{53.0} &34.7 & - & 86.4 \\
STG-KD~\cite{STGKD} & Vision &37.2 &27.3 &59.1 &44.6 &45.8 &34.3 &71.0 &86.0 \\
MGRMP~\cite{MGRMP} &Vision &37.4 &27.0 &58.8 &42.3 &-&-&-&- \\
SGN~\cite{SGN} &Vision &39.6 &27.6 &59.6 &45.2 &48.2 &34.2 &69.8 &84.6\\
SGN~\cite{SGN} &Vision + Motion &40.8 &28.3 &60.8 &49.5 &52.8 &35.5 &72.9 &94.3\\
HMN~\cite{ye2022hierarchical} &Vision + Motion &40.9 &27.3 &60.6 &46.6 &51.5 &34.4 &71.8 &88.3 \\
RSFD (AR-B)~\cite{rsfd} &Vision &42.1 &{29.1} &61.2 &51.1 &49.2 &35.3 &72.1 &91.4\\
UNiVL~\cite{luo2020univl} &Vision + Motion &{42.2} &28.8 &61.2 &49.9 &- &- &- &-\\
% \multicolumn{1}{l}{\rowcolor{Gray}{\textit{Our teacher model}}} & Vision &41.1 &29.0 &61.6 &51.3 &50.2 &37.3 &73.4 &96.9 \\
{\textcolor{gray}{Our teacher model}} &\textcolor{gray}{Vision} &\textcolor{gray}{41.1} &\textcolor{gray}{29.0} &\textcolor{gray}{61.6} &\textcolor{gray}{51.3} &\textcolor{gray}{50.2} &\textcolor{gray}{37.3} &\textcolor{gray}{73.4} &\textcolor{gray}{96.9} \\
\midrule
\multicolumn{4}{l}
{\textbf{\textit{Coded measurement based methods}}} & \\
{Our baseline} & Coded measurement & 24.7 &21.7 &52.0 &16.8 &25.5 &23.4 &51.8 &33.7 \\
{SnapCap} & Coded measurement &\textbf{42.2} &\textbf{29.1} &\textbf{62.0} &\textbf{52.2} &51.7 &\textbf{36.5} &\textbf{73.5} &\textbf{94.7} \\
\bottomrule
\end{tabular}
\vspace{-2mm}
\caption{Evaluation results of different compared methods on MSRVTT~\cite{msrvtt} and MSVD~\cite{msvd} datasets. For a fair comparison, we only include the results whose inputs are only video frames and features are extracted with 2D models (Vision) or 3D models (Motion). For our baseline, we adopt the same network structure as in SnapCap. }
\label{Tab:main table}
\vspace{-4mm}
\end{table*}

\section{Experiments}
\label{sec: 4}

In this section, we conduct experiments and report results to demonstrate the effectiveness of our proposed framework. We first detail some experimental settings including the datasets, compared methods, evaluation metrics, and devices. 
Then, we comprehensively evaluate the performance of our framework on both simulated coded measurements and real data. Finally, some ablation experiments are carried on to verify the roles of different components.
Note that in all tables, we highlight the best results in boldface.

\subsection{Experimental Settings}

\noindent \textbf{Datasets:} We conduct experiments on \textbf{MSRVTT}~\cite{msrvtt} and \textbf{MSVD}~\cite{msvd}, two extensively used video captioning datasets. Specifically, the MSRVTT dataset consists of 10K video clips with 20 captions per video, which are separated into 6,513 training samples, 497 for validation, and 2,990 for testing following previous works~\cite{chen2021motion, ryu2021semantic, ye2022hierarchical, Gu_2023_CVPR}. For the MSVD dataset, we separate it into 1,200 training videos, 100 for validation, and 670 for testing, respectively, following previous works~\cite{chen2021motion, ryu2021semantic, ye2022hierarchical, Gu_2023_CVPR}.

\noindent \textbf{Evaluation Metrics:}
Following previous VC works~\cite{chen2021motion, ryu2021semantic, ye2022hierarchical, Gu_2023_CVPR}, we use BLEU@4~\cite{papineni2002bleu} (B), METEOR~\cite{denkowski2014meteor} (M), ROUGE~\cite{lin2004rouge} (R) and CIDEr~\cite{vedantam2015cider} (C) as the evaluation metrics using the public tool\footnote{\url{https://github.com/salaniz/pycocoevalcap}}. 

\noindent \textbf{Measurement Simulation:} Considering that no public benchmarks have been introduced to evaluate our methods for now, we propose to synthesize the coded measurement on MSRVTT and MSVD. Specifically, for a given scene, a measurement is generated by compressing and integrating every $B$ high-speed frames using the coding masks $\{\Cmat_{k}\}_{k=1}^B$, as defined by Eq.~\eqref{eq: sci1}. 
Hence, in our work, no other large pre-training datasets have been employed.

\begin{figure*}[]
\centering
\includegraphics[width=1.0\textwidth]{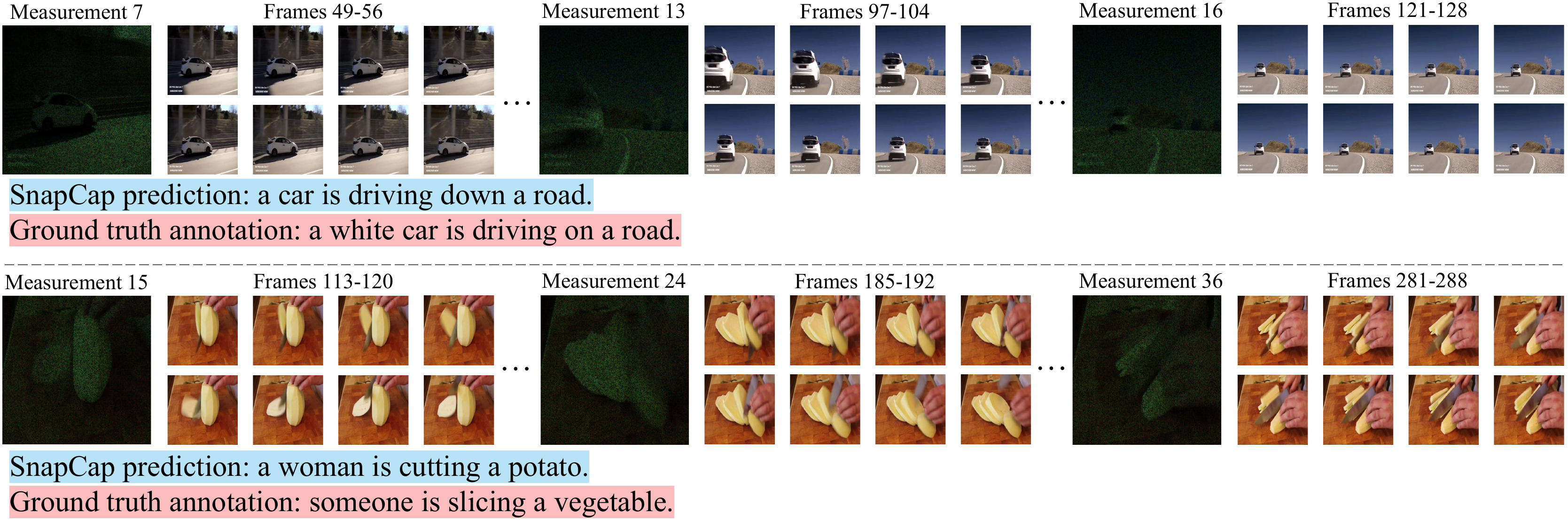}
 \vspace{-6mm}
\caption{Qualitative results on MSRVTT~\cite{msrvtt} (top row) and MSVD~\cite{msvd} (bottom row). We exhibit the compressed measurement, predicted caption by our SnapCap, and the ground truth. For a better understanding, we also show the ground truth video frames.}
\label{fig: caption results}
 \vspace{-4mm}
\end{figure*}

\noindent \textbf{Implementation Details:} All experiments are conducted on a workstation with 16 Intel i7 CPUs @ 2.50GHz and an NVIDIA Geforce RTX 3090 GPU. Similar to previous works~\cite{rsfd}, each video is segmented as $T=8$ clips, where we generate $T=8$ measurements with the mask ratio $B=8$ as the model input. We apply GPT-like Transformer~\cite{gpt2} as our caption generator. 
% For teacher model, we apply the CLIP ViT-B/16 as the vision encoder and our student model 
In Sec.~\ref{sec: loss}, we propose to optimize the parameters of $f(\cdot,\cdot)$ and $g(\cdot)$ firstly through the regularization in Eq.~\eqref{eq: recon loss} with AdamW optimizer~\cite{adamw} and the initial learning rate $3\times10^{-4}$ in the beginning 10 epochs and decaying to $1\times10^{-6}$ in the remaining 20 epochs. Secondly, to better extract meaningful and useful visual features for generating captions, we use AdamW optimizer~\cite{adamw} with learning rate $3\times10^{-4}$ for 30 epochs for the student model and the caption generator through loss function in Eq.~\eqref{eq: total loss}, where the coefficients $\alpha$ and $\beta$ are both set to 0.001. More details are in the Appendix. 

% During the first stage training, the parameters UNet are optimized through the AdamW optimizer~\cite{adamw} with initial learning rate $3\times10^{-4}$ in the beginning 10 epochs and decaying to $1\times10^{-6}$ in the remaining 20 epochs. 
% While in the second training stage, the parameters of modules are also updated through the AdamW optimizer~\cite{adamw} for 30 epochs. The loss coefficients $\alpha$ and $\beta$ are both set to 0.001. More details can be found in Supplementary Materials (SM). 

\subsection{Comparison with VC Methods}

To validate the effectiveness of our model, we conduct comparisons with SOTA video-based captioning methods on both MSRVTT and MSVD datasets. It should be noticed that, given video frames, most of SOTA methods employ one or more of spatial, motion, detection characteristics and others, \textit{e.g.}, external knowledge graph, audio transcripts, to generate captions, which takes more time to inference and consumes more storage. Here, for a fair comparison, we mainly compare our model with vision features-based algorithms. The quantitative results are listed in Table.~\ref{Tab:main table}. 
We also adopt a baseline model, which keeps the same architecture as our \textbf{SnapCap} and is only supervised by caption loss $\mathcal{L}_{cap}$ as in Eq.~\eqref{eq: cap loss}. During inference, $T$ coded measurements are directly input to the baseline to generate descriptions. 
For our SnapCap and baseline model, the pretrained CLIP ViT\footnote{Here, we load the official model CLIP ViT-B/16 from \url{https://github.com/openai/CLIP}}~\cite{CLIP} is employed as our teacher model and the weights of student model are initialized with CLIP. 

From Table.~\ref{Tab:main table}, it can be notably found the baseline model achieves much worse results, where during the training and inference of our baseline, we observe the severe over-fitting problem. Therefore it is rather difficult to obtain meaningful features directly from the coded measurement, as also observed in previous works~\cite{kumawat2022action, dong2023retrieving}. 
Equipped with the knowledge distillation strategy from video to measurement, our SnapCap demonstrates highly competitive performance compared to other video-based methods, and even the pre-trained large multi-modality model, UniVL~\cite{luo2020univl}. 
In Fig.~\ref{fig: caption results}, we visualize the coded measurement, video frames, predicted descriptions by our SnapCap as well as the ground truth. More qualitative results are presented in the Appendix.

\begin{table}
\centering
\begin{tabular}{ccccccc}
\toprule
\multirow{1}{*}{Settings} & $\mathcal{L}_{reg}$ & $\mathcal{L}_{dis}$ & B$\uparrow$ & R$\uparrow$ & M$\uparrow$ & C$\uparrow$ \\
\midrule
{Baseline} & & & 24.7 &21.7 &52.0 &16.8 \\
{a} &$\checkmark$ & & 32.1 &22.6 &55.6 &29.3 \\
{b} & &$\checkmark$ & 33.0 &24.9 &57.0 &31.6 \\
{SnapCap} &$\checkmark$  &$\checkmark$ &\textbf{42.2} &\textbf{29.1} &\textbf{62.0} &\textbf{52.2} \\
\bottomrule
\end{tabular}
\vspace{-2mm}
\caption{Ablation results on MSRVTT dataset~\cite{msrvtt}. ``$\checkmark$'' means we add the corresponding objective functions into the baseline.}
\vspace{-6mm}
\label{Tab: ablation}
\end{table}

\begin{table*}[ht]
\centering
\begin{tabular}{cccccccccc}
\toprule
\multirow{2}{*}{} & \multirow{2}{*}{Methods} 
&\multirow{2}{*}{Peak Memory} &\multicolumn{3}{c}{Inference time (ms)} &\multicolumn{4}{c}{MSRVTT} \\
\cmidrule(r){4-6} \cmidrule(r){7-10}
& & & Reconstruction & Caption &Total & B$\uparrow$ & R$\uparrow$ & M$\uparrow$ & C$\uparrow$ \\
\midrule
\multirow{5}{*}{Two-stage}
&BIRNAT~\cite{cheng2020birnat} &6.0GB &456 &485 &941 &38.4 &27.0 &59.7 &45.9 \\
&PnP-FFDNet~\cite{pnpffd} &6.3GB &6,011 &476 &6,486 &36.1 &26.6 &58.9 &40.8 \\
&PnP-FastDVDNet~\cite{pnpdvd} &6.3GB &10,300 &452 &10,752 &36.8 &26.5 &59.2 &42.4 \\
&STFormer~\cite{stformer} &17.0GB &825 &573 &1,398 &39.7 &28.2 &60.3 &48.8 \\
&EfficientSCI~\cite{wang2023efficientsci} &12.8GB &618 &462 &1,080 &39.3 &28.0 &60.6 &48.8 \\   
% &UNet &5.4GB &1000 &200 &1200 &24.7 &21.7 &52.0 &16.8 \\
\midrule
\multirow{2}{*}{One-stage} &Our Baseline &\textbf{5.6GB} &- &287 &287 &24.7 &21.7 &52.0 &16.8 \\
&SnapCap &6.1GB &- &\textbf{281} &\textbf{281} &\textbf{42.2} &\textbf{29.1} &\textbf{62.0} &\textbf{52.2} \\
\bottomrule
\end{tabular}
\vspace{-2mm}
\caption{Comparison of the complexity of different strategies, where for two-stage methods, we reconstruct videos first and then perform captioning using the teacher model trained caption generator from Table.~\ref{Tab:main table}. For all methods, we input $T=8$ measurements per video to the model and run on the same 3090 GPU. }
\vspace{-6mm}
\label{Tab: two-stage}
\end{table*}

\begin{table}
\centering
\begin{tabular}{ccccccc}
\toprule
$T$ & \makecell[c]{Memory\\(GB)} & \makecell[c]{Inference\\time (ms)} & B$\uparrow$ & R$\uparrow$ & M$\uparrow$ & C$\uparrow$ \\
\midrule
{2} &5.4 &\textbf{227} &40.0 &28.1 &60.6 &48.8 \\
{4} &5.8 &251 &41.1 &28.6 &61.4 &50.7 \\
{8} &6.1 &281 &\textbf{42.2} &\textbf{29.1} &\textbf{62.0} &\textbf{52.2} \\
{12} &7.3 &314 &41.8 &{29.0} &61.8 &51.9 \\
\bottomrule
\end{tabular}
\vspace{-2mm}
\caption{Ablation experiments on the number of measurements $T$ per video on MSRVTT dataset~\cite{msrvtt}, where the compression ratio $B$ is set to 8 for all $T$.}
\vspace{-2mm}
\label{Tab: meas number}
\end{table}

\subsection{Ablation Study}

\subsubsection{Comparison with two-stage methods}
Given the coded measurement, a straightforward and intuitive manner for captioning is to reconstruct frames first and then perform captioning. 
% However, such a two-stage strategy potentially lacks privacy protection on the one hand.
% \textit{i.e.}, in some surveillance cases where obtaining a precise reconstruction of the scene is not needed but we are still interested in the semantic information.
However, such a two-stage strategy typically consumes more time and computational resources, which poses a tricky dilemma in resource-limited occasions. 
In this part, we conduct experiments to demonstrate the superiority of our \textit{reconstruction-free compression-free compressive learning schema} in terms of inference speed, memory consumption, and captioning quality on the same 3090 GPU.
For two-stage methods, we load pre-trained neural networks BIRNAT~\cite{cheng2020birnat}, EfficientSCI~\cite{wang2023efficientsci} and STFormer~\cite{stformer}, and plug-and-play methods PnP-FFDNet~\cite{pnpffd}, PnP-FastDVDNet~\cite{pnpdvd} to perform reconstruction first with $T=8$ input measurements. Then a trained captioning model\footnote{Here, we use the CLIP ViT-B/16 and corresponding language decoder module, \textit{aka}, teacher model in Table.~\ref{Tab:main table}.} is applied to generate descriptions. The results on MSRVTT~\cite{msrvtt} are listed in the Table.~\ref{Tab: two-stage}, where the inference time is averaged over the whole testing set with the batch size 1. 
It can be clearly found our method has significant advantages in terms of the inference speed and it also achieves the best captioning performance among these methods. 

\begin{figure}[]
\centering
\includegraphics[width=1\columnwidth]{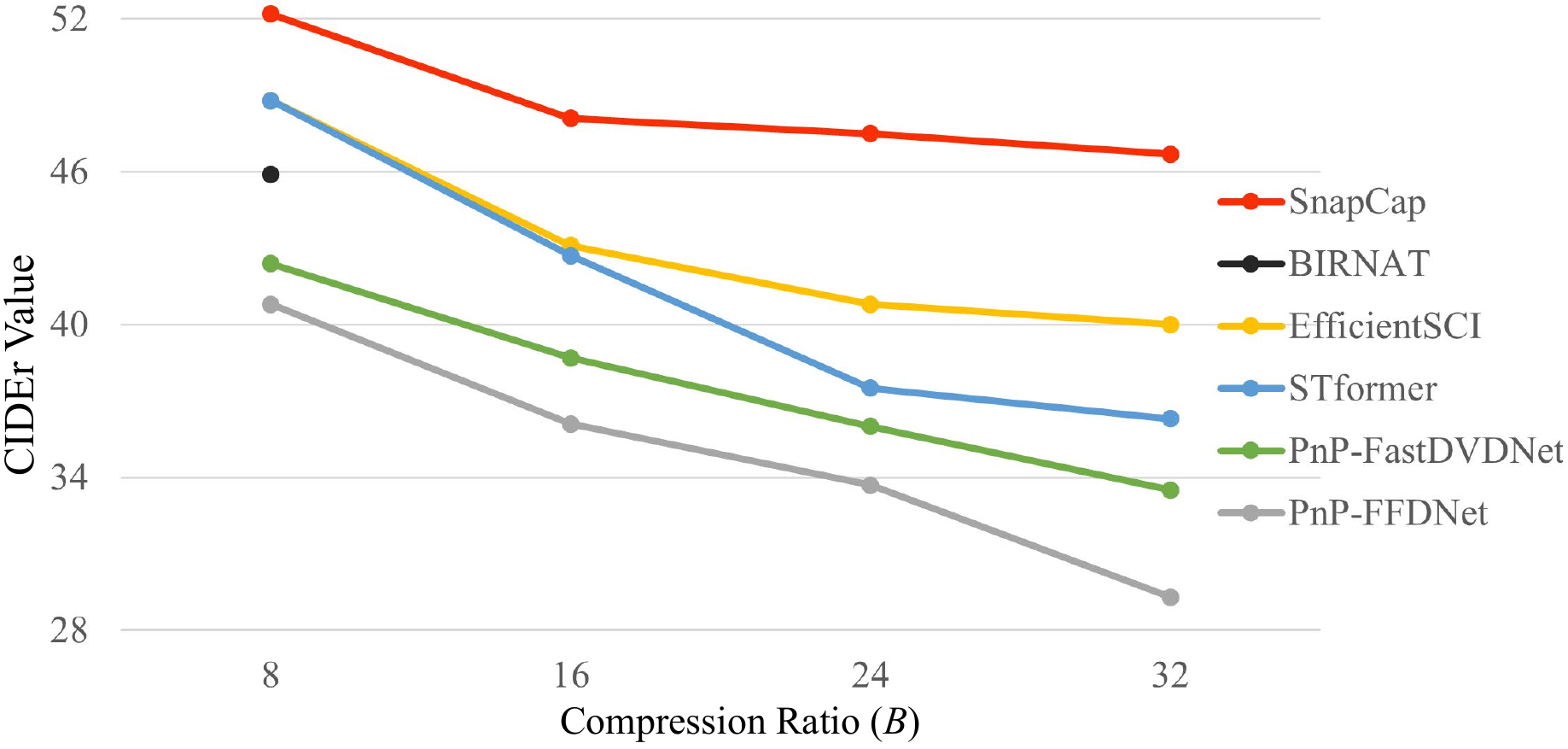}
 \vspace{-5mm}
\caption{Captioning quality (in terms of CIDEr value) comparison of different methods with different compression ratios $B$ as in Table.~\ref{Tab: two-stage}. }
\label{fig: compression ratio}
\vspace{-6mm}
\end{figure}

Further, given the fact that in video CS systems such as CACTI~\cite{cacti}, 
% and subsequent reconstruction tasks~\cite{wang2023efficientsci, cheng2020birnat}, 
the compression ratio $B$ plays a determined factor in the quality of recovered videos for software decoders. 
Usually, the smaller $B$, the better reconstruction quality, leading to better captioning performances. 
To evaluate the robustness of SnapCap, we conduct experiments with different $B$ and report the CIDEr values~\cite{vedantam2015cider} in Fig.~\ref{fig: compression ratio}. 
From the figure, it can be notably found that our SnapCap shows the least performance degradation as $B$ increases.

\subsubsection{Effects of regularization and distillation}
\label{sec: regularization ablation}

In Sec.~\ref{sec: encoder}, we introduce a novel VC pipeline which takes the compressed measurement and the masks as input to derive language-related vision features for captioning. 
During training, we propose to optimize the encoder $f(\cdot, \cdot)$ and decoder $g(\cdot)$ through the $\mathcal{L}_{reg}$ loss firstly and then update the parameters of $f(\cdot, \cdot)$,  $S()$ and $\mathrm{Proj}(\cdot)$ under the guidance of teacher model secondly as well as the captioning loss. 
To demonstrate the effectiveness of transferring knowledge strategy through the regularization manner and the direct feature map matching schema, we conduct experiments by adding the $\mathcal{L}_{reg}$ and $\mathcal{L}_{dis}$ step by step on the baseline model. The numerical results on MSRVTT~\cite{msrvtt} are reported in Table.~\ref{Tab: ablation}. It can be remarkably found that both two knowledge transferring strategies take effect to extract meaningful and language-related vision features for captioning. 
% For setting a, we add the $\mathcal{L}_{recon}$ loss during the first-stage training, and such design can help the student model extract meaningful features after the UNet encoder. Further, the introduction of knowledge transfer from CLIP~\cite{CLIP} effectively captures the semantic information from the coded measurement to generate high-quality captions. 
% Both of $\mathcal{L}_{recon}$ and $\mathcal{L}_{dis}$ contribute much to the performance of our model.
% where it can be clearly found that our knowledge distillation design . 

\subsubsection{Impact of the number of measurements}

In previous VC works~\cite{lin2022swinbert}, different numbers of frames for each video may lead to different performances, which corresponds to the number of measurements $T$ per video in our model.
Hence, in this part, we conduct experiments with varied number of measurements $T$ to verify the robustness of our framework, where the compression ratio is fixed to 8, and the results are listed in Table.~\ref{Tab: meas number}.
It can be seen that as the number of measurement increases, with more information involved, our SnapCap achieves consistent performance improvements.

\begin{figure}[]
\centering
\includegraphics[width=1\columnwidth]{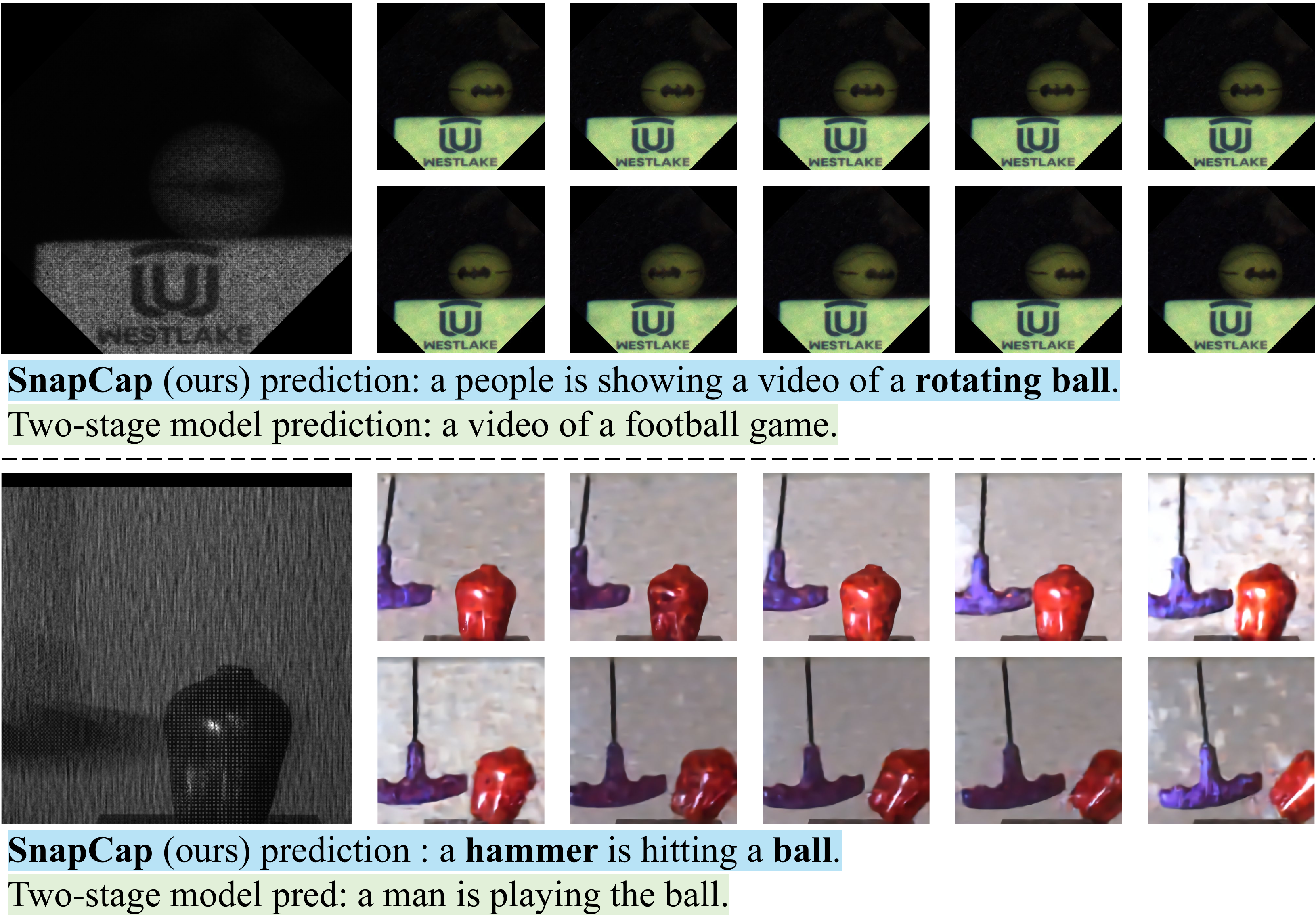}
 \vspace{-6mm}
\caption{Comparison of captioning results (our model prediction and two-stage model prediction) on two real data. The top row is about $\mathrm{Ball}$~$\mathrm{Rotate}$ from~\cite{stformer} and the bottom is about $\mathrm{Hammer}$ from~\cite{desci}. For better understanding, we also plot the reconstructed results of STFormer~\cite{stformer} and BIRNAT~\cite{cheng2020birnat}. }
\label{fig: real data}
 \vspace{-6mm}
\end{figure}

\subsection{Real Datasets}
Except for simulation data, we also apply our framework to the real data captured by the CACTI system. To be more specific, we test our model on two public real snapshot compressive data, $\mathrm{Ball}$ $\mathrm{Rotate}$~\cite{stformer} and $\mathrm{Hammer}$~\cite{desci}, which are captured by~\cite{yuan2014low}. 
The coded measurement and our predicted caption are presented in Fig.~\ref{fig: real data}, where the reconstructions obtained by STFormer~\cite{stformer} and BIRNAT~\cite{cheng2020birnat} are also exhibited for reference. It can be clearly and notably noticed that our proposed VC pipeline is able to describe the scene accurately in language. 

\section{Conclusion and Future Work}
In this paper, to achieve efficient video captioning without the software based compression nor reconstruction, we propose a novel end-to-end framework to generate captions directly from the compressed measurement. Specifically, we employ the knowledge distillation strategy through a pre-trained large vision-language CLIP to transfer the knowledge from the video domain to the measurment domain. 
Compared to two-stage methods, our proposed SnapCap is able to describe the scene efficiently and accurately. We also verify the feasibility of our model in real data.
However, our SnapCap can only be used for captioning up to now. 
In the future, we will explore to extend our framework to other video-related tasks with on other datasets.

% Also, combining with the hardware system, we intend to regulate the encoding process with the feedback strategy.
% However, some issues still exist that . 

% \section{Discussions and Future Work}

% In Table.~\ref{Tab:main table}, it can be notably found that the baseline model achieves rather bad performances. During the training process, the over-fitting issue has been observed and we attribute it to the lack of training samples. Besides, we conduct experiments on only video captioning task, where we believe that language is the most concise and efficient expression of semantics. Hence, in the future, \textit{i)}. we plan to incorporate a large video-text set for training; \textit{ii)}, we will explore more video-related tasks in the measurement domain; \textit{iii)}. combining with the hardware system, we intend to regulate the CACTI encoding process with feedback strategy.

\clearpage
{
    \small
    \bibliographystyle{ieeenat_fullname}
    \bibliography{main}

\begin{thebibliography}{80}
\providecommand{\natexlab}[1]{#1}
\providecommand{\url}[1]{\texttt{#1}}
\expandafter\ifx\csname urlstyle\endcsname\relax
  \providecommand{\doi}[1]{doi: #1}\else
  \providecommand{\doi}{doi: \begingroup \urlstyle{rm}\Url}\fi

\bibitem[Abdar et~al.(2023)Abdar, Kollati, Kuraparthi, Pourpanah, McDuff,
  Ghavamzadeh, Yan, Mohamed, Khosravi, Cambria, and Porikli]{abdar2023review}
Moloud Abdar, Meenakshi Kollati, Swaraja Kuraparthi, Farhad Pourpanah, Daniel
  McDuff, Mohammad Ghavamzadeh, Shuicheng Yan, Abduallah Mohamed, Abbas
  Khosravi, Erik Cambria, and Fatih Porikli.
\newblock A review of deep learning for video captioning, 2023.

\bibitem[Bain et~al.(2021)Bain, Nagrani, Varol, and Zisserman]{bain2021frozen}
Max Bain, Arsha Nagrani, G{\"u}l Varol, and Andrew Zisserman.
\newblock Frozen in time: A joint video and image encoder for end-to-end
  retrieval.
\newblock In \emph{Proceedings of the IEEE/CVF International Conference on
  Computer Vision}, pages 1728--1738, 2021.

\bibitem[Barbastathis et~al.(2019)Barbastathis, Ozcan, and Situ]{ci2}
George Barbastathis, Aydogan Ozcan, and Guohai Situ.
\newblock On the use of deep learning for computational imaging.
\newblock \emph{Optica}, 6\penalty0 (8):\penalty0 921--943, 2019.

\bibitem[Chang et~al.(2023)Chang, Wang, Xu, Chen, Yang, and
  Zhao]{Chang_2023_ICCV}
Jiahao Chang, Shuo Wang, Hai-Ming Xu, Zehui Chen, Chenhongyi Yang, and Feng
  Zhao.
\newblock Detrdistill: A universal knowledge distillation framework for
  detr-families.
\newblock In \emph{Proceedings of the IEEE/CVF International Conference on
  Computer Vision (ICCV)}, pages 6898--6908, 2023.

\bibitem[Chen and Dolan(2011)]{msvd}
David Chen and William~B Dolan.
\newblock Collecting highly parallel data for paraphrase evaluation.
\newblock In \emph{Proceedings of the 49th annual meeting of the association
  for computational linguistics: human language technologies}, pages 190--200,
  2011.

\bibitem[Chen et~al.(2023)Chen, Zhu, Qian, Ghanem, Yan, Zhu, Xiao, Culatana,
  and Elhoseiny]{Chen_2023_ICCV}
Jun Chen, Deyao Zhu, Guocheng Qian, Bernard Ghanem, Zhicheng Yan, Chenchen Zhu,
  Fanyi Xiao, Sean~Chang Culatana, and Mohamed Elhoseiny.
\newblock Exploring open-vocabulary semantic segmentation from clip vision
  encoder distillation only.
\newblock In \emph{Proceedings of the IEEE/CVF International Conference on
  Computer Vision (ICCV)}, pages 699--710, 2023.

\bibitem[Chen and Jiang(2019)]{MGSA}
Shaoxiang Chen and Yu-Gang Jiang.
\newblock Motion guided spatial attention for video captioning.
\newblock In \emph{Proceedings of the AAAI conference on artificial
  intelligence}, pages 8191--8198, 2019.

\bibitem[Chen and Jiang(2021{\natexlab{a}})]{MGRMP}
Shaoxiang Chen and Yu-Gang Jiang.
\newblock Motion guided region message passing for video captioning.
\newblock In \emph{Proceedings of the IEEE/CVF International Conference on
  Computer Vision}, pages 1543--1552, 2021{\natexlab{a}}.

\bibitem[Chen and Jiang(2021{\natexlab{b}})]{chen2021motion}
Shaoxiang Chen and Yu-Gang Jiang.
\newblock Motion guided region message passing for video captioning.
\newblock In \emph{Proceedings of the IEEE/CVF International Conference on
  Computer Vision}, pages 1543--1552, 2021{\natexlab{b}}.

\bibitem[Cheng et~al.(2020)Cheng, Lu, Wang, Zhang, Chen, Meng, and
  Yuan]{cheng2020birnat}
Ziheng Cheng, Ruiying Lu, Zhengjue Wang, Hao Zhang, Bo Chen, Ziyi Meng, and Xin
  Yuan.
\newblock Birnat: Bidirectional recurrent neural networks with adversarial
  training for video snapshot compressive imaging.
\newblock In \emph{European Conference on Computer Vision}, pages 258--275.
  Springer, 2020.

\bibitem[Cheng et~al.(2021)Cheng, Chen, Liu, Zhang, Lu, Wang, and Yuan]{RevSCI}
Ziheng Cheng, Bo Chen, Guanliang Liu, Hao Zhang, Ruiying Lu, Zhengjue Wang, and
  Xin Yuan.
\newblock Memory-efficient network for large-scale video compressive sensing.
\newblock In \emph{Proceedings of the IEEE/CVF Conference on Computer Vision
  and Pattern Recognition}, pages 16246--16255, 2021.

\bibitem[Cho et~al.(2020)Cho, Kwak, Yoon, and Kim]{cho2020speech}
Won~Ik Cho, Donghyun Kwak, Ji~Won Yoon, and Nam~Soo Kim.
\newblock Speech to text adaptation: Towards an efficient cross-modal
  distillation.
\newblock \emph{arXiv preprint arXiv:2005.08213}, 2020.

\bibitem[Denkowski and Lavie(2014)]{denkowski2014meteor}
Michael Denkowski and Alon Lavie.
\newblock Meteor universal: Language specific translation evaluation for any
  target language.
\newblock In \emph{Proceedings of the ninth workshop on statistical machine
  translation}, pages 376--380, 2014.

\bibitem[Devlin et~al.(2018)Devlin, Chang, Lee, and Toutanova]{devlin2018bert}
Jacob Devlin, Ming-Wei Chang, Kenton Lee, and Kristina Toutanova.
\newblock Bert: Pre-training of deep bidirectional transformers for language
  understanding.
\newblock \emph{arXiv preprint arXiv:1810.04805}, 2018.

\bibitem[Dong et~al.(2023)Dong, Guo, Yang, Cheng, Suo, and
  Dai]{dong2023retrieving}
Kaiming Dong, Yuchen Guo, Runzhao Yang, Yuxiao Cheng, Jinli Suo, and Qionghai
  Dai.
\newblock Retrieving object motions from coded shutter snapshot in dark
  environment.
\newblock \emph{IEEE Transactions on Image Processing}, 2023.

\bibitem[Ge et~al.(2021)Ge, Zhang, Choi, Cheung, Zhao, Zhu, Wang, Zhao, and
  Li]{ge2021self}
Yixiao Ge, Xiao Zhang, Ching~Lam Choi, Ka~Chun Cheung, Peipei Zhao, Feng Zhu,
  Xiaogang Wang, Rui Zhao, and Hongsheng Li.
\newblock Self-distillation with batch knowledge ensembling improves imagenet
  classification.
\newblock \emph{arXiv preprint arXiv:2104.13298}, 2021.

\bibitem[Gu et~al.(2021)Gu, Lin, Kuo, and Cui]{gu2021zero}
Xiuye Gu, Tsung-Yi Lin, Weicheng Kuo, and Yin Cui.
\newblock Zero-shot detection via vision and language knowledge distillation.
\newblock \emph{arXiv preprint arXiv:2104.13921}, 2\penalty0 (3):\penalty0 4,
  2021.

\bibitem[Gu et~al.(2023{\natexlab{a}})Gu, Chen, Wang, Zhang, Luo, and
  Wen]{Gu_2023_CVPR}
Xin Gu, Guang Chen, Yufei Wang, Libo Zhang, Tiejian Luo, and Longyin Wen.
\newblock Text with knowledge graph augmented transformer for video captioning.
\newblock In \emph{Proceedings of the IEEE/CVF Conference on Computer Vision
  and Pattern Recognition (CVPR)}, pages 18941--18951, 2023{\natexlab{a}}.

\bibitem[Gu et~al.(2023{\natexlab{b}})Gu, Chen, Wang, Zhang, Luo, and
  Wen]{textkg}
Xin Gu, Guang Chen, Yufei Wang, Libo Zhang, Tiejian Luo, and Longyin Wen.
\newblock Text with knowledge graph augmented transformer for video captioning.
\newblock In \emph{Proceedings of the IEEE/CVF Conference on Computer Vision
  and Pattern Recognition}, pages 18941--18951, 2023{\natexlab{b}}.

\bibitem[Gupta et~al.(2016)Gupta, Hoffman, and Malik]{gupta2016cross}
Saurabh Gupta, Judy Hoffman, and Jitendra Malik.
\newblock Cross modal distillation for supervision transfer.
\newblock In \emph{Proceedings of the IEEE conference on computer vision and
  pattern recognition}, pages 2827--2836, 2016.

\bibitem[He et~al.(2016)He, Zhang, Ren, and Sun]{resnet}
Kaiming He, Xiangyu Zhang, Shaoqing Ren, and Jian Sun.
\newblock Deep residual learning for image recognition.
\newblock In \emph{Proceedings of the IEEE conference on computer vision and
  pattern recognition}, pages 770--778, 2016.

\bibitem[He et~al.(2022)He, Chen, Xie, Li, Doll{\'a}r, and
  Girshick]{he2022masked}
Kaiming He, Xinlei Chen, Saining Xie, Yanghao Li, Piotr Doll{\'a}r, and Ross
  Girshick.
\newblock Masked autoencoders are scalable vision learners.
\newblock In \emph{Proceedings of the IEEE/CVF conference on computer vision
  and pattern recognition}, pages 16000--16009, 2022.

\bibitem[Hu et~al.(2021)Hu, Huang, Chen, Yang, and Chen]{hu2021video}
Chengyang Hu, Honghao Huang, Minghua Chen, Sigang Yang, and Hongwei Chen.
\newblock Video object detection from one single image through opto-electronic
  neural network.
\newblock \emph{APL Photonics}, 6\penalty0 (4), 2021.

\bibitem[Jalali and Yuan(2019)]{JalaliY19}
Shirin Jalali and Xin Yuan.
\newblock Snapshot compressed sensing: Performance bounds and algorithms.
\newblock \emph{{IEEE} Trans. Inf. Theory}, 65\penalty0 (12):\penalty0
  8005--8024, 2019.

\bibitem[Jiao et~al.(2019)Jiao, Yin, Shang, Jiang, Chen, Li, Wang, and
  Liu]{jiao2019tinybert}
Xiaoqi Jiao, Yichun Yin, Lifeng Shang, Xin Jiang, Xiao Chen, Linlin Li, Fang
  Wang, and Qun Liu.
\newblock Tinybert: Distilling bert for natural language understanding.
\newblock \emph{arXiv preprint arXiv:1909.10351}, 2019.

\bibitem[Kim et~al.(2020)Kim, Jung, and Lee]{kim2020few}
Geonuk Kim, Hong-Gyu Jung, and Seong-Whan Lee.
\newblock Few-shot object detection via knowledge transfer.
\newblock In \emph{2020 IEEE International Conference on Systems, Man, and
  Cybernetics (SMC)}, pages 3564--3569. IEEE, 2020.

\bibitem[Kumawat et~al.(2022)Kumawat, Okawara, Yoshida, Nagahara, and
  Yagi]{kumawat2022action}
Sudhakar Kumawat, Tadashi Okawara, Michitaka Yoshida, Hajime Nagahara, and
  Yasushi Yagi.
\newblock Action recognition from a single coded image.
\newblock \emph{IEEE Transactions on Pattern Analysis and Machine
  Intelligence}, 45\penalty0 (4):\penalty0 4109--4121, 2022.

\bibitem[Li et~al.(2020{\natexlab{a}})Li, Lin, Ding, Liu, Zhu, and
  Han]{li2020gan}
Muyang Li, Ji Lin, Yaoyao Ding, Zhijian Liu, Jun-Yan Zhu, and Song Han.
\newblock Gan compression: Efficient architectures for interactive conditional
  gans.
\newblock In \emph{Proceedings of the IEEE/CVF conference on computer vision
  and pattern recognition}, pages 5284--5294, 2020{\natexlab{a}}.

\bibitem[Li et~al.(2022)Li, Lin, Wang, Fei, Shao, and Ji]{li2022learning}
Shaojie Li, Mingbao Lin, Yan Wang, Chao Fei, Ling Shao, and Rongrong Ji.
\newblock Learning efficient gans for image translation via differentiable
  masks and co-attention distillation.
\newblock \emph{IEEE Transactions on Multimedia}, 2022.

\bibitem[Li et~al.(2020{\natexlab{b}})Li, Qi, Gulve, Wei, Genov, Kutulakos, and
  Heidrich]{li2020end}
Yuqi Li, Miao Qi, Rahul Gulve, Mian Wei, Roman Genov, Kiriakos~N Kutulakos, and
  Wolfgang Heidrich.
\newblock End-to-end video compressive sensing using anderson-accelerated
  unrolled networks.
\newblock In \emph{2020 IEEE international conference on computational
  photography (ICCP)}, pages 1--12. IEEE, 2020{\natexlab{b}}.

\bibitem[Lin(2004)]{lin2004rouge}
Chin-Yew Lin.
\newblock Rouge: A package for automatic evaluation of summaries.
\newblock In \emph{Text summarization branches out}, pages 74--81, 2004.

\bibitem[Lin et~al.(2022)Lin, Li, Lin, Ahmed, Gan, Liu, Lu, and
  Wang]{lin2022swinbert}
Kevin Lin, Linjie Li, Chung-Ching Lin, Faisal Ahmed, Zhe Gan, Zicheng Liu,
  Yumao Lu, and Lijuan Wang.
\newblock Swinbert: End-to-end transformers with sparse attention for video
  captioning.
\newblock In \emph{Proceedings of the IEEE/CVF Conference on Computer Vision
  and Pattern Recognition}, pages 17949--17958, 2022.

\bibitem[Liu et~al.(2018)Liu, Yuan, Suo, Brady, and Dai]{desci}
Yang Liu, Xin Yuan, Jinli Suo, David~J Brady, and Qionghai Dai.
\newblock Rank minimization for snapshot compressive imaging.
\newblock \emph{IEEE transactions on pattern analysis and machine
  intelligence}, 41\penalty0 (12):\penalty0 2990--3006, 2018.

\bibitem[Llull et~al.(2013)Llull, Liao, Yuan, Yang, Kittle, Carin, Sapiro, and
  Brady]{cacti}
Patrick Llull, Xuejun Liao, Xin Yuan, Jianbo Yang, David Kittle, Lawrence
  Carin, Guillermo Sapiro, and David~J Brady.
\newblock Coded aperture compressive temporal imaging.
\newblock \emph{Optics express}, 21\penalty0 (9):\penalty0 10526--10545, 2013.

\bibitem[Loshchilov and Hutter(2017)]{adamw}
Ilya Loshchilov and Frank Hutter.
\newblock Decoupled weight decay regularization.
\newblock \emph{arXiv preprint arXiv:1711.05101}, 2017.

\bibitem[Luo et~al.(2020)Luo, Ji, Shi, Huang, Duan, Li, Li, Bharti, and
  Zhou]{luo2020univl}
Huaishao Luo, Lei Ji, Botian Shi, Haoyang Huang, Nan Duan, Tianrui Li, Jason
  Li, Taroon Bharti, and Ming Zhou.
\newblock Univl: A unified video and language pre-training model for multimodal
  understanding and generation.
\newblock \emph{arXiv preprint arXiv:2002.06353}, 2020.

\bibitem[Ma et~al.(2019)Ma, Liu, Shou, and Yuan]{ma2019deep}
Jiawei Ma, Xiao-Yang Liu, Zheng Shou, and Xin Yuan.
\newblock Deep tensor admm-net for snapshot compressive imaging.
\newblock In \emph{Proceedings of the IEEE/CVF International Conference on
  Computer Vision}, pages 10223--10232, 2019.

\bibitem[Mait et~al.(2018)Mait, Euliss, and Athale]{ci1}
Joseph~N Mait, Gary~W Euliss, and Ravindra~A Athale.
\newblock Computational imaging.
\newblock \emph{Advances in Optics and Photonics}, 10\penalty0 (2):\penalty0
  409--483, 2018.

\bibitem[Miech et~al.(2019)Miech, Zhukov, Alayrac, Tapaswi, Laptev, and
  Sivic]{miech19howto100m}
Antoine Miech, Dimitri Zhukov, Jean-Baptiste Alayrac, Makarand Tapaswi, Ivan
  Laptev, and Josef Sivic.
\newblock How{T}o100{M}: {L}earning a {T}ext-{V}ideo {E}mbedding by {W}atching
  {H}undred {M}illion {N}arrated {V}ideo {C}lips.
\newblock In \emph{ICCV}, 2019.

\bibitem[Pan et~al.(2020)Pan, Cai, Huang, Lee, Gaidon, Adeli, and
  Niebles]{STGKD}
Boxiao Pan, Haoye Cai, De-An Huang, Kuan-Hui Lee, Adrien Gaidon, Ehsan Adeli,
  and Juan~Carlos Niebles.
\newblock Spatio-temporal graph for video captioning with knowledge
  distillation.
\newblock In \emph{Proceedings of the IEEE/CVF Conference on Computer Vision
  and Pattern Recognition}, pages 10870--10879, 2020.

\bibitem[Papineni et~al.(2002)Papineni, Roukos, Ward, and
  Zhu]{papineni2002bleu}
Kishore Papineni, Salim Roukos, Todd Ward, and Wei-Jing Zhu.
\newblock Bleu: a method for automatic evaluation of machine translation.
\newblock In \emph{Proceedings of the 40th annual meeting of the Association
  for Computational Linguistics}, pages 311--318, 2002.

\bibitem[Qiao et~al.(2020)Qiao, Meng, Ma, and Yuan]{qiao2020deep}
Mu Qiao, Ziyi Meng, Jiawei Ma, and Xin Yuan.
\newblock Deep learning for video compressive sensing.
\newblock \emph{Apl Photonics}, 5\penalty0 (3), 2020.

\bibitem[Radford et~al.(2019)Radford, Wu, Child, Luan, Amodei, Sutskever,
  et~al.]{gpt2}
Alec Radford, Jeffrey Wu, Rewon Child, David Luan, Dario Amodei, Ilya
  Sutskever, et~al.
\newblock Language models are unsupervised multitask learners.
\newblock \emph{OpenAI blog}, 1\penalty0 (8):\penalty0 9, 2019.

\bibitem[Radford et~al.(2021)Radford, Kim, Hallacy, Ramesh, Goh, Agarwal,
  Sastry, Askell, Mishkin, Clark, et~al.]{CLIP}
Alec Radford, Jong~Wook Kim, Chris Hallacy, Aditya Ramesh, Gabriel Goh,
  Sandhini Agarwal, Girish Sastry, Amanda Askell, Pamela Mishkin, Jack Clark,
  et~al.
\newblock Learning transferable visual models from natural language
  supervision.
\newblock In \emph{International conference on machine learning}, pages
  8748--8763. PMLR, 2021.

\bibitem[Ramanishka et~al.(2017)Ramanishka, Das, Zhang, and
  Saenko]{ramanishka2017top}
Vasili Ramanishka, Abir Das, Jianming Zhang, and Kate Saenko.
\newblock Top-down visual saliency guided by captions.
\newblock In \emph{Proceedings of the IEEE conference on computer vision and
  pattern recognition}, pages 7206--7215, 2017.

\bibitem[Ryu et~al.(2021{\natexlab{a}})Ryu, Kang, Kang, and Yoo]{SGN}
Hobin Ryu, Sunghun Kang, Haeyong Kang, and Chang~D Yoo.
\newblock Semantic grouping network for video captioning.
\newblock In \emph{proceedings of the AAAI Conference on Artificial
  Intelligence}, pages 2514--2522, 2021{\natexlab{a}}.

\bibitem[Ryu et~al.(2021{\natexlab{b}})Ryu, Kang, Kang, and
  Yoo]{ryu2021semantic}
Hobin Ryu, Sunghun Kang, Haeyong Kang, and Chang~D Yoo.
\newblock Semantic grouping network for video captioning.
\newblock In \emph{proceedings of the AAAI Conference on Artificial
  Intelligence}, pages 2514--2522, 2021{\natexlab{b}}.

\bibitem[Schuhmann et~al.(2021)Schuhmann, Vencu, Beaumont, Kaczmarczyk, Mullis,
  Katta, Coombes, Jitsev, and Komatsuzaki]{schuhmann2021laion}
Christoph Schuhmann, Richard Vencu, Romain Beaumont, Robert Kaczmarczyk,
  Clayton Mullis, Aarush Katta, Theo Coombes, Jenia Jitsev, and Aran
  Komatsuzaki.
\newblock Laion-400m: Open dataset of clip-filtered 400 million image-text
  pairs.
\newblock \emph{arXiv preprint arXiv:2111.02114}, 2021.

\bibitem[Seo et~al.(2022)Seo, Nagrani, Arnab, and Schmid]{mvgpt}
Paul~Hongsuck Seo, Arsha Nagrani, Anurag Arnab, and Cordelia Schmid.
\newblock End-to-end generative pretraining for multimodal video captioning.
\newblock In \emph{Proceedings of the IEEE/CVF Conference on Computer Vision
  and Pattern Recognition}, pages 17959--17968, 2022.

\bibitem[Szegedy et~al.(2016)Szegedy, Ioffe, Vanhoucke, and
  Alemi]{szegedy2016inceptionv4}
Christian Szegedy, Sergey Ioffe, Vincent Vanhoucke, and Alex Alemi.
\newblock Inception-v4, inception-resnet and the impact of residual connections
  on learning, 2016.

\bibitem[Tang et~al.(2021)Tang, Wang, Liu, Rao, Li, and
  Li]{tang2021clip4caption}
Mingkang Tang, Zhanyu Wang, Zhenhua Liu, Fengyun Rao, Dian Li, and Xiu Li.
\newblock Clip4caption: Clip for video caption.
\newblock In \emph{Proceedings of the 29th ACM International Conference on
  Multimedia}, pages 4858--4862, 2021.

\bibitem[Tewel et~al.(2022)Tewel, Shalev, Nadler, Schwartz, and
  Wolf]{tewel2022zero}
Yoad Tewel, Yoav Shalev, Roy Nadler, Idan Schwartz, and Lior Wolf.
\newblock Zero-shot video captioning with evolving pseudo-tokens.
\newblock \emph{arXiv preprint arXiv:2207.11100}, 2022.

\bibitem[Tran et~al.(2015)Tran, Bourdev, Fergus, Torresani, and Paluri]{C3D}
Du Tran, Lubomir Bourdev, Rob Fergus, Lorenzo Torresani, and Manohar Paluri.
\newblock Learning spatiotemporal features with 3d convolutional networks.
\newblock In \emph{Proceedings of the IEEE international conference on computer
  vision}, pages 4489--4497, 2015.

\bibitem[Vedantam et~al.(2015)Vedantam, Lawrence~Zitnick, and
  Parikh]{vedantam2015cider}
Ramakrishna Vedantam, C Lawrence~Zitnick, and Devi Parikh.
\newblock Cider: Consensus-based image description evaluation.
\newblock In \emph{Proceedings of the IEEE conference on computer vision and
  pattern recognition}, pages 4566--4575, 2015.

\bibitem[Wang et~al.(2018{\natexlab{a}})Wang, Ma, Zhang, and Liu]{recnet}
Bairui Wang, Lin Ma, Wei Zhang, and Wei Liu.
\newblock Reconstruction network for video captioning.
\newblock In \emph{Proceedings of the IEEE conference on computer vision and
  pattern recognition}, pages 7622--7631, 2018{\natexlab{a}}.

\bibitem[Wang et~al.(2022{\natexlab{a}})Wang, Yang, Hu, Li, Lin, Gan, Liu, Liu,
  and Wang]{wang2022git}
Jianfeng Wang, Zhengyuan Yang, Xiaowei Hu, Linjie Li, Kevin Lin, Zhe Gan,
  Zicheng Liu, Ce Liu, and Lijuan Wang.
\newblock Git: A generative image-to-text transformer for vision and language.
\newblock \emph{arXiv preprint arXiv:2205.14100}, 2022{\natexlab{a}}.

\bibitem[Wang et~al.(2022{\natexlab{b}})Wang, Cao, Zhong, and Yuan]{stformer}
Lishun Wang, Miao Cao, Yong Zhong, and Xin Yuan.
\newblock Spatial-temporal transformer for video snapshot compressive imaging.
\newblock \emph{IEEE Transactions on Pattern Analysis and Machine
  Intelligence}, 2022{\natexlab{b}}.

\bibitem[Wang et~al.(2023)Wang, Cao, and Yuan]{wang2023efficientsci}
Lishun Wang, Miao Cao, and Xin Yuan.
\newblock Efficientsci: Densely connected network with space-time factorization
  for large-scale video snapshot compressive imaging.
\newblock In \emph{Proceedings of the IEEE/CVF Conference on Computer Vision
  and Pattern Recognition}, pages 18477--18486, 2023.

\bibitem[Wang et~al.(2019)Wang, Yuan, Zhang, and Feng]{wang2019distilling}
Tao Wang, Li Yuan, Xiaopeng Zhang, and Jiashi Feng.
\newblock Distilling object detectors with fine-grained feature imitation.
\newblock In \emph{Proceedings of the IEEE/CVF Conference on Computer Vision
  and Pattern Recognition}, pages 4933--4942, 2019.

\bibitem[Wang et~al.(2018{\natexlab{b}})Wang, Zhang, Sun, and
  Qi]{wang2018kdgan}
Xiaojie Wang, Rui Zhang, Yu Sun, and Jianzhong Qi.
\newblock Kdgan: Knowledge distillation with generative adversarial networks.
\newblock \emph{Advances in neural information processing systems}, 31,
  2018{\natexlab{b}}.

\bibitem[Wang et~al.(2021)Wang, Zhang, Cheng, Chen, and Yuan]{wang2021metasci}
Zhengjue Wang, Hao Zhang, Ziheng Cheng, Bo Chen, and Xin Yuan.
\newblock Metasci: Scalable and adaptive reconstruction for video compressive
  sensing.
\newblock In \emph{Proceedings of the IEEE/CVF Conference on Computer Vision
  and Pattern Recognition}, pages 2083--2092, 2021.

\bibitem[Wu et~al.(2023{\natexlab{a}})Wu, Peng, Zhou, Xiao, Liu, Yuan, Xuan,
  Valenzuela, Chen, Wang, Chao, and Hu]{Wu_2023_ICCV}
Kan Wu, Houwen Peng, Zhenghong Zhou, Bin Xiao, Mengchen Liu, Lu Yuan, Hong
  Xuan, Michael Valenzuela, Xi~(Stephen) Chen, Xinggang Wang, Hongyang Chao,
  and Han Hu.
\newblock Tinyclip: Clip distillation via affinity mimicking and weight
  inheritance.
\newblock In \emph{Proceedings of the IEEE/CVF International Conference on
  Computer Vision (ICCV)}, pages 21970--21980, 2023{\natexlab{a}}.

\bibitem[Wu et~al.(2021)Wu, Zhang, and Mou]{wu2021dense}
Zhuoyuan Wu, Jian Zhang, and Chong Mou.
\newblock Dense deep unfolding network with 3d-cnn prior for snapshot
  compressive imaging.
\newblock \emph{arXiv preprint arXiv:2109.06548}, 2021.

\bibitem[Wu et~al.(2023{\natexlab{b}})Wu, Yang, Su, and Yuan]{wu2023adaptive}
Zongliang Wu, Chengshuai Yang, Xiongfei Su, and Xin Yuan.
\newblock Adaptive deep pnp algorithm for video snapshot compressive imaging.
\newblock \emph{International Journal of Computer Vision}, pages 1--18,
  2023{\natexlab{b}}.

\bibitem[Xie et~al.(2018)Xie, Sun, Huang, Tu, and Murphy]{S3D}
Saining Xie, Chen Sun, Jonathan Huang, Zhuowen Tu, and Kevin Murphy.
\newblock Rethinking spatiotemporal feature learning: Speed-accuracy trade-offs
  in video classification.
\newblock In \emph{Proceedings of the European conference on computer vision
  (ECCV)}, pages 305--321, 2018.

\bibitem[Xu et~al.(2023)Xu, Ye, Yan, Shi, Ye, Xu, Li, Bi, Qian, Wang,
  et~al.]{mplug2}
Haiyang Xu, Qinghao Ye, Ming Yan, Yaya Shi, Jiabo Ye, Yuanhong Xu, Chenliang
  Li, Bin Bi, Qi Qian, Wei Wang, et~al.
\newblock mplug-2: A modularized multi-modal foundation model across text,
  image and video.
\newblock \emph{arXiv preprint arXiv:2302.00402}, 2023.

\bibitem[Xu et~al.(2016)Xu, Mei, Yao, and Rui]{msrvtt}
Jun Xu, Tao Mei, Ting Yao, and Yong Rui.
\newblock Msr-vtt: A large video description dataset for bridging video and
  language.
\newblock In \emph{Proceedings of the IEEE conference on computer vision and
  pattern recognition}, pages 5288--5296, 2016.

\bibitem[Xu et~al.(2020)Xu, Rui, Li, and Gu]{xu2020feature}
Kunran Xu, Lai Rui, Yishi Li, and Lin Gu.
\newblock Feature normalized knowledge distillation for image classification.
\newblock In \emph{European conference on computer vision}, pages 664--680.
  Springer, 2020.

\bibitem[Yang et~al.(2021)Yang, Zou, Liu, and Zhang]{yang2021NACF}
Bang Yang, Yuexian Zou, Fenglin Liu, and Can Zhang.
\newblock Non-autoregressive coarse-to-fine video captioning.
\newblock In \emph{Proceedings of the AAAI Conference on Artificial
  Intelligence}, pages 3119--3127, 2021.

\bibitem[Ye et~al.(2022)Ye, Li, Qi, Wang, Huang, and Yang]{ye2022hierarchical}
Hanhua Ye, Guorong Li, Yuankai Qi, Shuhui Wang, Qingming Huang, and Ming-Hsuan
  Yang.
\newblock Hierarchical modular network for video captioning.
\newblock In \emph{Proceedings of the IEEE/CVF Conference on Computer Vision
  and Pattern Recognition}, pages 17939--17948, 2022.

\bibitem[Yu et~al.(2019)Yu, Yazici, Liu, Weijer, Cheng, and
  Ramisa]{yu2019learning}
Lu Yu, Vacit~Oguz Yazici, Xialei Liu, Joost van~de Weijer, Yongmei Cheng, and
  Arnau Ramisa.
\newblock Learning metrics from teachers: Compact networks for image embedding.
\newblock In \emph{Proceedings of the IEEE/CVF Conference on Computer Vision
  and Pattern Recognition}, pages 2907--2916, 2019.

\bibitem[Yuan(2016)]{gaptv}
Xin Yuan.
\newblock Generalized alternating projection based total variation minimization
  for compressive sensing.
\newblock In \emph{2016 IEEE International conference on image processing
  (ICIP)}, pages 2539--2543. IEEE, 2016.

\bibitem[Yuan et~al.(2014)Yuan, Llull, Liao, Yang, Brady, Sapiro, and
  Carin]{yuan2014low}
Xin Yuan, Patrick Llull, Xuejun Liao, Jianbo Yang, David~J Brady, Guillermo
  Sapiro, and Lawrence Carin.
\newblock Low-cost compressive sensing for color video and depth.
\newblock In \emph{Proceedings of the IEEE Conference on Computer Vision and
  Pattern Recognition}, pages 3318--3325, 2014.

\bibitem[Yuan et~al.(2020{\natexlab{a}})Yuan, Liu, Suo, and Dai]{pnpffd}
Xin Yuan, Yang Liu, Jinli Suo, and Qionghai Dai.
\newblock Plug-and-play algorithms for large-scale snapshot compressive
  imaging.
\newblock In \emph{Proceedings of the IEEE/CVF Conference on Computer Vision
  and Pattern Recognition}, pages 1447--1457, 2020{\natexlab{a}}.

\bibitem[Yuan et~al.(2020{\natexlab{b}})Yuan, Liu, Suo, and Dai]{yuan2020plug}
Xin Yuan, Yang Liu, Jinli Suo, and Qionghai Dai.
\newblock Plug-and-play algorithms for large-scale snapshot compressive
  imaging.
\newblock In \emph{Proceedings of the IEEE/CVF Conference on Computer Vision
  and Pattern Recognition}, pages 1447--1457, 2020{\natexlab{b}}.

\bibitem[Yuan et~al.(2021{\natexlab{a}})Yuan, Brady, and Katsaggelos]{sci1}
Xin Yuan, David~J Brady, and Aggelos~K Katsaggelos.
\newblock Snapshot compressive imaging: Theory, algorithms, and applications.
\newblock \emph{IEEE Signal Processing Magazine}, 38\penalty0 (2):\penalty0
  65--88, 2021{\natexlab{a}}.

\bibitem[Yuan et~al.(2021{\natexlab{b}})Yuan, Liu, Suo, Durand, and
  Dai]{pnpdvd}
Xin Yuan, Yang Liu, Jinli Suo, Fredo Durand, and Qionghai Dai.
\newblock Plug-and-play algorithms for video snapshot compressive imaging.
\newblock \emph{IEEE Transactions on Pattern Analysis and Machine
  Intelligence}, 44\penalty0 (10):\penalty0 7093--7111, 2021{\natexlab{b}}.

\bibitem[Zhang et~al.(2023)Zhang, Guo, Jiao, Zhang, and
  Han]{zhang2023efficient}
Tianlu Zhang, Hongyuan Guo, Qiang Jiao, Qiang Zhang, and Jungong Han.
\newblock Efficient rgb-t tracking via cross-modality distillation.
\newblock In \emph{Proceedings of the IEEE/CVF Conference on Computer Vision
  and Pattern Recognition}, pages 5404--5413, 2023.

\bibitem[Zhang et~al.(2022)Zhang, Zhang, Yuan, Zheng, Su, Suo, Brady, and
  Dai]{zhang2022compressive}
Zhihong Zhang, Bo Zhang, Xin Yuan, Siming Zheng, Xiongfei Su, Jinli Suo,
  David~J Brady, and Qionghai Dai.
\newblock From compressive sampling to compressive tasking: Retrieving
  semantics in compressed domain with low bandwidth.
\newblock \emph{PhotoniX}, 3\penalty0 (1):\penalty0 1--22, 2022.

\bibitem[Zhong et~al.(2023)Zhong, Li, Chen, Jiang, Chen, and Ye]{rsfd}
Xian Zhong, Zipeng Li, Shuqin Chen, Kui Jiang, Chen Chen, and Mang Ye.
\newblock Refined semantic enhancement towards frequency diffusion for video
  captioning.
\newblock In \emph{Thirty-Seventh {AAAI} Conference on Artificial Intelligence,
  {AAAI} 2023, Thirty-Fifth Conference on Innovative Applications of Artificial
  Intelligence, {IAAI} 2023, Thirteenth Symposium on Educational Advances in
  Artificial Intelligence, {EAAI} 2023, Washington, DC, USA, February 7-14,
  2023}, pages 3724--3732. {AAAI} Press, 2023.

\end{thebibliography}
}


\clearpage 

In this Appendix, we mainly include the network architectures, learning and inference algorithms, the parameter settings and some visualizations. 

\section{Network Architecture}

\begin{figure}[h]
\centering
\includegraphics[width=1.\columnwidth]{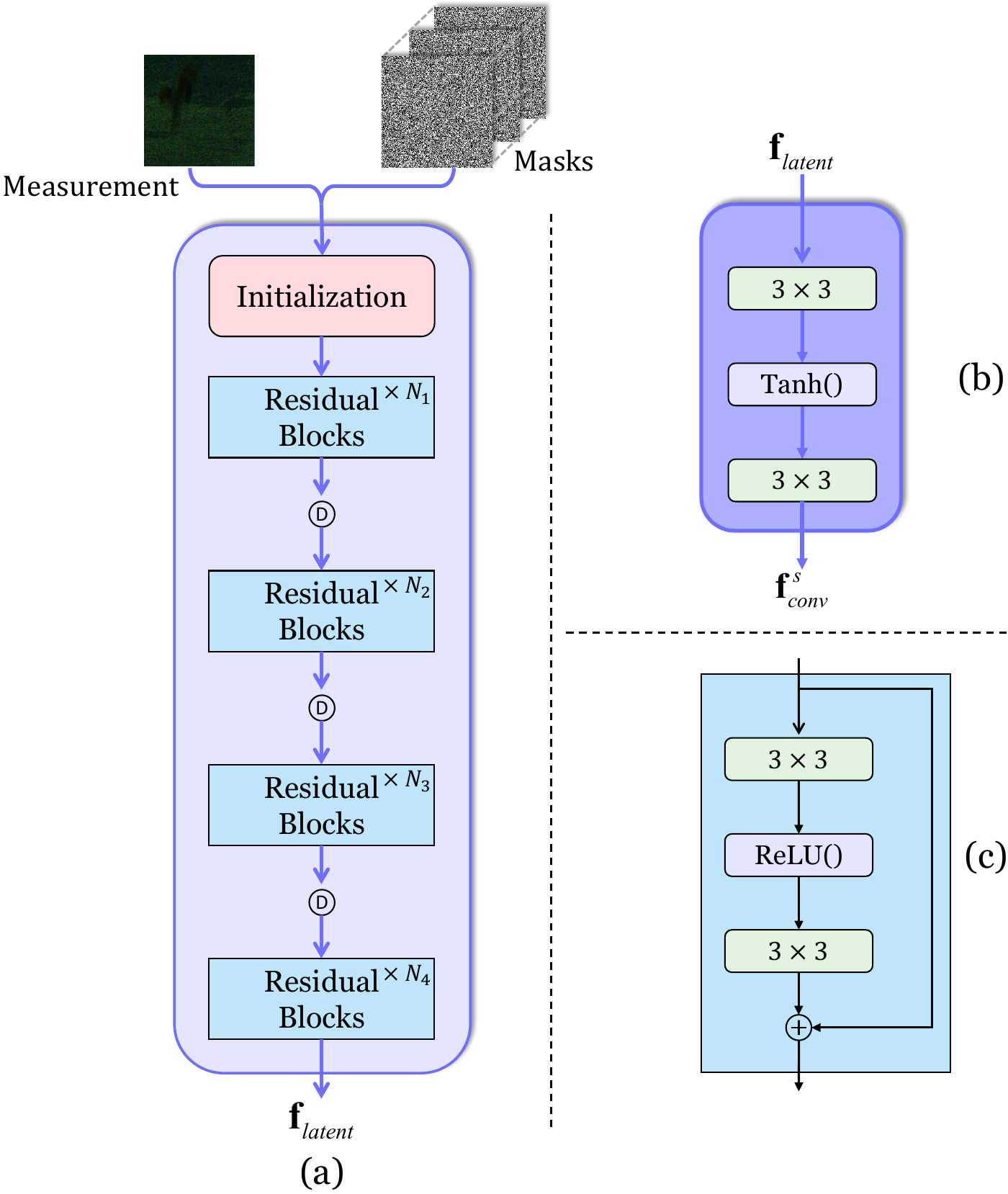}
% \vspace{-6mm}
\caption{Detailed architectures of (a) encoder $f(\cdot, \cdot)$; (b) $\mathrm{Conv2}(\cdot)$; and (c) the residual block used in (a). ``D'' means the down-sampling operation to reduce the spatial resolution, ``$3\times3$'' denotes the convolutional operation with the kernel size $3$, $\mathrm{Tanh}()$ and $\mathrm{ReLU}()$ are both activation functions. }
\label{fig: encoder}
% \vspace{-10mm}
\end{figure}

In Sec.~\ref{sec: encoder}, we propose an encoder $f(\cdot, \cdot)$ which consists of multiple residual blocks and a two-layer flexible convolutional operation $\mathrm{Conv2}(\cdot)$ to extract feature maps from the measurement domain. In Fig.~\ref{fig: encoder}, we present the detailed architecture of the encoder $f(\cdot, \cdot)$ in (a) and $\mathrm{Conv2}(\cdot)$ in (b). 

Following previous video snapshot compressive sensing works, we introduce an ``Initialization'' module into our architecture, as shown in Fig.~\ref{fig: encoder}.
Specifically, the ``Initialization'' module fuse the information of the coded measurement $\Ymat$ and the masks $\Cmat$ via:
\begin{align}
\bar{\Ymat}&=\Ymat \oslash \sum_{k=1}^{B} \Cmat_{k},\\
\Xmat_{e}&=\bar{\Ymat} \odot \Cmat + \bar{\Ymat},
\end{align}
where $\bar{\Ymat}$ is the normalized measurement, $\oslash$ and $\odot$ are the element-wise division and element-wise multiplication, respectively. In this manner, a coarse estimate of the video frames can be obtained with more information compared to the measurement. The residual block numbers $N_1, N_2, N_3$ and $N_4$ in Fig.~\ref{fig: encoder} are set as $4, 6, 6$ and $6$, respectively.

Then, to enable an efficient regularization training, we design a decoder $g(\cdot)$ mentioned in Eq.~\eqref{eq: conv loss}, which also consists of multiple residual blocks as the encoder $f(\cdot, \cdot)$. 

\section{Learning and Inference Algorithms}

\begin{algorithm}
\caption{Inference stage}
\label{alg: test}
\KwData{Coded measurement $\Ymat$ and masks $\Cmat$.}
\KwIn{Trained models encoder $f(\cdot, \cdot)$, student model $S(\cdot)$, projector $\mathrm{Proj}(\cdot)$, and a pre-trained Language decoder $\mathrm{Dec}(\cdot)$.}
\KwOut{Predicted captions.}

{
Input the measurement $\Ymat$ and masks $\{\Cmat_{k}\}_{k=1}^{B}$ to the the student model $S(\cdot)$ to get the latent representation $\boldsymbol{\mathrm{f}}_{latent}$ as in Eq. (\textcolor{blue}{3}) and visual embedding $\boldsymbol{\mathrm{f}}^s$ as in Eq. (\textcolor{blue}{6}), respectively;

% Input the same measurement $\Ymat$ and masks $\{\Cmat_{k}\}_{k=1}^{B}$ to to the student model $S(\cdot)$ to obtain the visual embedding $\boldsymbol{\mathrm{f}}^s$ as in Eq. (\textcolor{blue}{6});
 
Input visual embedding $\boldsymbol{\mathrm{f}}^{s}$ to a projector $\mathrm{Proj}(\cdot)$ to obtain $\boldsymbol{\mathrm{t}}$ as in Eq. (\textcolor{blue}{11});

Generate the predicted caption word-by-word through the language decoder $\mathrm{Dec}()$ as in Eq. (\textcolor{blue}{14}).
}
\end{algorithm}

In Sec.~\ref{sec: methodology} of our main paper, the student model is optimized through the proposed knowledge distillation module under the guidance of CLIP model on the video domain. 
Detailed learning and inference algorithms are presented in Alg.~\ref{alg: train} and Alg.~\ref{alg: test}, respectively.

\begin{algorithm}
\caption{Learning stage}
\label{alg: train}
\KwData{Distribution over video frames: $p(\mathcal{T})$}
\KwIn{Masks, $\{\Cmat_{k}\}_{k=1}^{B}$; Loss coefficients $\alpha$ and $\beta$; A pre-trained Language Encoder $\mathrm{PLM}(\cdot)$; A pre-trained Language decoder $\mathrm{Dec}(\cdot)$}
\KwOut{Trained parameters for encoder $f(\cdot,\cdot)$, student model $S(\cdot)$, projector $\mathrm{Proj}(\cdot)$.}
\For{\textnormal{epoch}=\textnormal{1, 2, ..., 30}}
{Randomly sample a video $\mathcal{T}_{i} \sim p(\mathcal{T})$, and sample $B$ video frames $\{\Xmat_{k}\}_{k=1}^B$ from $\mathcal{T}_{i}$;

Simulate the coded measurement $\Ymat$ with masks $\{\Cmat_{k}\}_{k=1}^{B}$ as in Eq. (\textcolor{blue}{1});

Input the measurement $\Ymat$ and the masks $\{\Cmat_{k}\}_{k=1}^{B}$ to the encoder $f(\cdot,\cdot)$, and then decoder $g(\cdot)$ to obtain $\hat{\Xmat}$ as in Eq. (\textcolor{blue}{10});

Update the parameters of encoder $f(\cdot,\cdot)$ and decoder $g(\cdot)$ through the regularization loss in Eq. (\textcolor{blue}{10}).
}

\For{\textnormal{epoch}=\textnormal{1, 2, ..., 30}}
{Randomly sample a video $\mathcal{T}_{i}$, and generate the coded measurement $\Ymat$ with masks $\Cmat$ as in Eq. (\textcolor{blue}{1}) from $B$ video frames $\Xmat$;

% Input the measurement $\Ymat$ and masks $\{\Cmat_{k}\}_{k=1}^{B}$ to the encoder $f(\cdot,\cdot)$ to get the latent representation $\boldsymbol{\mathrm{f}}_{latent}$ as in Eq. (\textcolor{blue}{3});

Input the measurement $\Ymat$ and masks $\{\Cmat_{k}\}_{k=1}^{B}$ to the student model $S(\cdot)$ to obtain the feature maps $\boldsymbol{\mathrm{f}}_{conv}^s$ as in Eq. (\textcolor{blue}{4}) and the visual embedding $\boldsymbol{\mathrm{f}}^s$ as in Eq. (\textcolor{blue}{6}), respectively;

Input the video frames $\Xmat$ to the teacher model $T(\cdot)$ to obtain the feature maps $\boldsymbol{\mathrm{f}}_{conv}^t$ as in Eq. (\textcolor{blue}{2}) and the visual embedding $\boldsymbol{\mathrm{f}}^t$ as in Eq. (\textcolor{blue}{5});

Compute the distillation loss $\mathcal{L}_{dis}$ as in Eq. (\textcolor{blue}{7}) to Eq. (\textcolor{blue}{9});

Input visual embedding $\boldsymbol{\mathrm{f}}^s$ to a projector $\mathrm{Proj}(\cdot)$ to obtain $\boldsymbol{\mathrm{t}}$ as in Eq. (\textcolor{blue}{11});

Input the ground truth annotation to the $\mathrm{PLM}(\cdot)$ and generate the predicted caption word-by-word as in Eq. (\textcolor{blue}{12}) to Eq. (\textcolor{blue}{14});

Update the parameters of encoder $f(\cdot, \cdot)$, student model $S(\cdot)$, and the projector $\mathrm{Proj}(\cdot)$.
}
\end{algorithm}

\section{Hyperparameter Settings}

Table.~\ref{tab: recon} and Table.~\ref{tab: total} list the main hyperparameters used in our experiments.

\begin{table}[h]
    \centering
    \begin{tabular}{l|c} 
        \toprule[1pt]
        \textbf{Configs} & \textbf{Values} \\
        \hline
        Input resolution &$224\times224$ \\
        Optimizer & AdamW \\
        Base learning rate & 3e-4\\
        Weight decay & 1e-4\\
        Optimizer momentum & $\beta_1, \beta_2=0.9,0.999$ \\
        Learning rate schedule & \makecell{Cosine Annealing\\ Restart Cyclic} \\
        Warmup epochs & 10 \\
        Training epochs &30 \\
        Batch size & 32\\
        \bottomrule[1pt]
    \end{tabular}
    \caption{Hyperparameter settings in the regularization training stage. }
    \label{tab: recon}
\end{table}

\begin{table}[h]
    \centering
    \begin{tabular}{l|c} 
        \toprule[1pt]
        \textbf{Configs} & \textbf{Values} \\
        \hline
        Input resolution &$224\times224$ \\
        Feature map dim & $c,h,w=768,14,14$ \\
        Vision embedding dim &$d=512$ \\
        Text embedding dim &$D=1024$ \\
        Optimizer & AdamW \\
        Learning rate ($\mathrm{Conv2(\cdot)}$) & 1e-6\\
        Learning rate ($\mathrm{S(\cdot)}$) & 1e-6\\
        Learning rate ($\mathrm{Proj(\cdot)}$) & 3e-4\\
        Weight decay & 1e-4\\
        Optimizer momentum & $\beta_1, \beta_2=0.9,0.999$ \\
        Batch size & 32\\
        Training epochs &30 \\
        \bottomrule[1pt]
    \end{tabular}
    \caption{Hyperparameter settings in the distillation and caption training stage. }
    \label{tab: total}
\end{table}

\section{More Results}

Except for the qualitative results presented in the Fig.~\ref{fig: caption results}in our main paper, more visualization results on the MSRVTT and MSVD datasets are provided below.

\begin{figure*}[]
\centering
\includegraphics[width=1.0\textwidth]{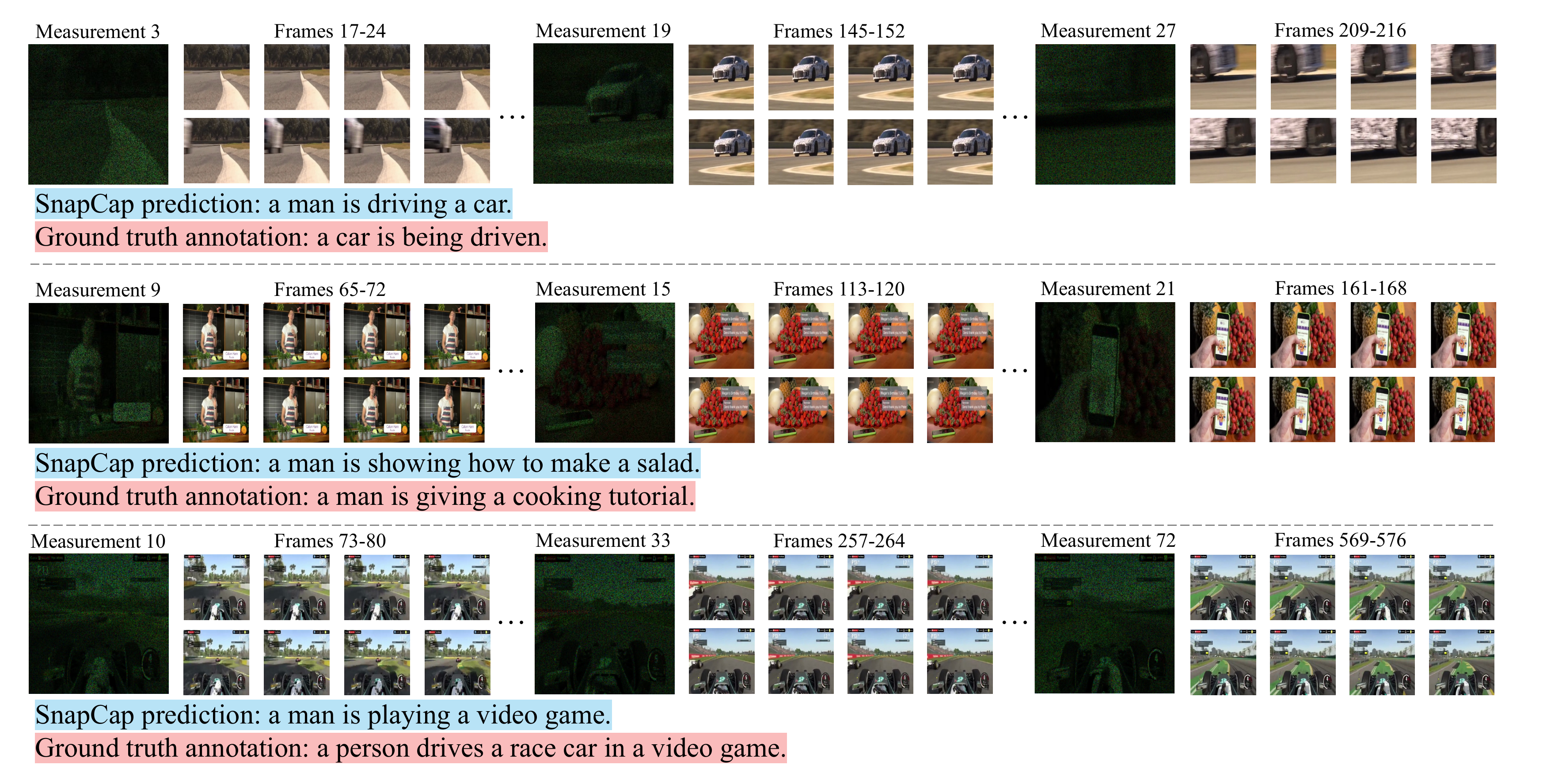}
 % \vspace{-6mm}
\caption{Qualitative results from three different videos on MSRVTT dataset. We exhibit the compressed measurement, predicted caption by our SnapCap, and the ground truth. For a better understanding, we also show the ground truth video frames.}
\label{fig: msrvtt results}
 % \vspace{-4mm}
\end{figure*}

\begin{figure*}[]
\centering
\includegraphics[width=1.0\textwidth]{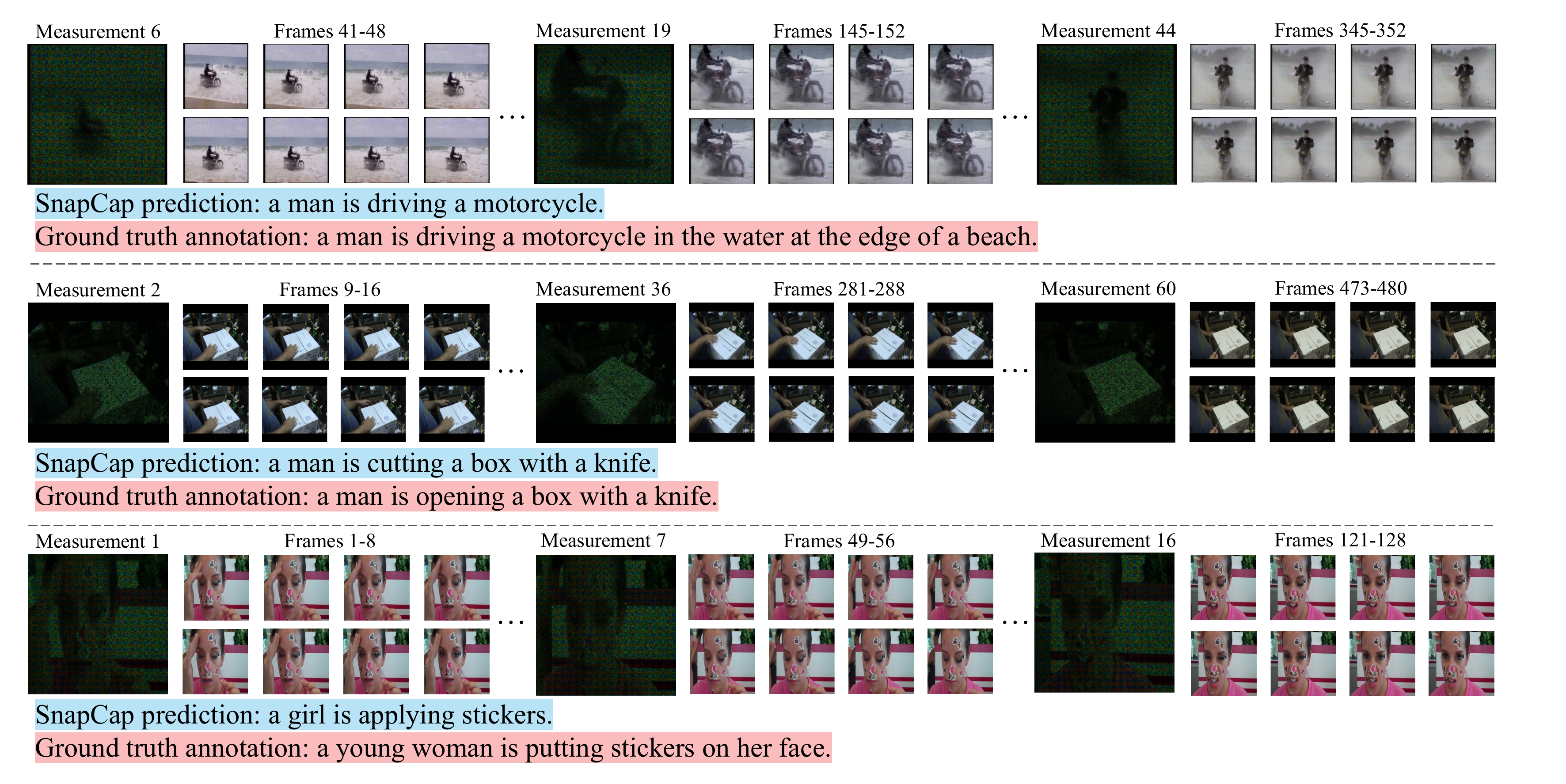}
 % \vspace{-6mm}
\caption{Qualitative results from three different videos on MSVD dataset. We exhibit the compressed measurement, predicted caption by our SnapCap, and the ground truth. For a better understanding, we also show the ground truth video frames.}
\label{fig: msvd results}
 % \vspace{-4mm}
\end{figure*}

\end{document}